\documentclass[pdflatex,sn-mathphys-ay,iicol]{sn-jnl}

\usepackage{mathtools}
\newcommand{\R}{\mathbb{R}}
\renewcommand{\O}{\mathcal{O}}
\newcommand{\N}{\mathcal{N}}
\newcommand{\diag}[1]{\mathrm{diag}\left(#1\right)}
\newcommand{\tr}[1]{\mathrm{tr}\left(#1\right)}
\newcommand{\lrp}[1]{\left(#1\right)}
\newcommand{\lrb}[1]{\left[#1\right]}
\newcommand{\lrs}[1]{\left\{#1\right\}}

\newcommand{\Sym}{\mathrm{Sym}}
\newcommand{\T}{^\top}
\DeclareMathOperator*{\argmin}{arg\,min}
\DeclareMathOperator*{\argmax}{arg\,max}

\usepackage{graphicx}%
\usepackage{multirow}%
\usepackage{amsmath,amssymb,amsfonts}%
\usepackage{amsthm}%
\usepackage{mathrsfs}%
\usepackage[title]{appendix}%
\usepackage{xcolor}%
\usepackage{textcomp}%
\usepackage{manyfoot}%
\usepackage{booktabs}%
\usepackage{algorithm}%
\usepackage{algorithmicx}%
\usepackage{algpseudocode}%
\usepackage{listings}%

\theoremstyle{thmstyleone}%
\newtheorem{theorem}{Theorem}
\newtheorem{proposition}[theorem]{Proposition}%

\theoremstyle{thmstyletwo}%
\newtheorem{remark}{Remark}%

\theoremstyle{thmstylethree}%
\newtheorem{definition}{Definition}%

\begin{document}

\title[Parsimonious Gaussian mixture models with piecewise-constant eigenvalue profiles]{Parsimonious Gaussian mixture models with piecewise-constant eigenvalue profiles}


\author*[1]{\fnm{Tom} \sur{Szwagier}}\email{tom.szwagier@inria.fr}

\author[2]{\fnm{Pierre-Alexandre} \sur{Mattei}}\email{pierre-alexandre.mattei@inria.fr}

\author[2]{\fnm{Charles} \sur{Bouveyron}}\email{charles.bouveyron@inria.fr}

\author[1]{\fnm{Xavier} \sur{Pennec}}\email{xavier.pennec@inria.fr}

\affil*[1]{\orgname{Université Côte d'Azur}, \orgdiv{INRIA, Epione Project Team}, \orgaddress{\street{2004 Rte des Lucioles}, \postcode{06902}, \city{Sophia Antipolis}, \country{France}}}

\affil[2]{\orgname{Université Côte d'Azur}, \orgdiv{INRIA, CNRS, LJAD, Maasai Project Team}, \orgaddress{\street{2004 Rte des Lucioles}, \postcode{06902}, \city{Sophia Antipolis}, \country{France}}}


\abstract{
    Gaussian mixture models (GMMs) are ubiquitous in statistical learning, particularly for unsupervised problems. While full GMMs suffer from the overparameterization of their covariance matrices in high-dimensional spaces, spherical GMMs (with isotropic covariance matrices) certainly lack flexibility to fit certain anisotropic distributions. Connecting these two extremes, we introduce a new family of parsimonious GMMs with piecewise-constant covariance eigenvalue profiles. These extend several low-rank models like the celebrated mixtures of probabilistic principal component analyzers (MPPCA), by enabling any possible sequence of eigenvalue multiplicities. If the latter are prespecified, then we can naturally derive an expectation--maximization (EM) algorithm to learn the mixture parameters.  Otherwise, to address the notoriously-challenging issue of jointly learning the mixture parameters and hyperparameters, we propose a componentwise penalized EM algorithm, whose monotonicity is proven. We show the superior likelihood--parsimony tradeoffs achieved by our models on a variety of unsupervised experiments: density fitting, clustering and single-image denoising.
}



\maketitle

\section{Introduction}\label{sec:intro}
Gaussian mixture models (GMMs)~\citep{mclachlan_finite_2000} are ubiquitous in statistical learning. Their probabilistic nature gives them many advantages against classical machine learning algorithms such as random forests~\citep{breiman_random_2001}, support vector machines~\citep{boser_training_1992}, K-Means~\citep{lloyd_least_1982}, HDBSCAN~\citep{mcinnes_hdbscan_2017} and spectral clustering~\citep{ng_spectral_2001}. They notably enable uncertainty estimates, likelihood-based model selection, generative modeling and they can handle missing data.
Gaussian-based models show good performances in low-to-moderate dimensions and are therefore heavily used in basic tasks such as classification or clustering.
However, the overparameterization of their covariance matrices---$p(p+1)/2$ parameters in dimension $p$---is a clear issue in high dimensions, where data often lack. Hence, there is a need for simpler models.

Many works have attempted to overcome the high-dimensional issues of GMMs (see for instance the Chapter 8 of~\cite{bouveyron_model-based_2019} for a review). A widely-used heuristic is to reduce the dimension of the data---e.g. via principal component analysis~\citep{jolliffe_principal_2002}---before fitting the mixture model. This practical method may however obviously discard some important information for the subsequent task; it may for instance mix some clusters that were well separated in the original space~\citep{chang_using_1983}.
Other methods---referred to as \textit{parsimonious}---consider GMMs in the full data space but with fewer parameters. The early works of~\cite{banfield_model-based_1993,celeux_gaussian_1995,bensmail_regularized_1996} consider GMMs with two simple parsimony-inducing principles: parameter sharing (e.g. constraining the mixture components to have the same covariance matrix) and covariance constraints (e.g. imposing diagonal or scalar covariance matrices). However, the former may not reduce enough the number of parameters in high dimensions, while the latter may reduce it too much. As a consequence, many intermediate parsimonious models later appeared, with low-rank~\citep{ghahramani_em_1996,mclachlan_modelling_2003,mcnicholas_parsimonious_2008}, sparse~\citep{fop_model-based_2019,cappozzo_sparse_2025}, Markov random field~\citep{finder_effective_2022}, ultrametric~\citep{cavicchia_parsimonious_2024}, or structured priors~\citep{heaps_structured_2024} for the covariance matrices, with an important impact~\citep{fraley_model-based_2002,scrucca_mclust_2016}.

Among the parsimonious GMM family, the celebrated mixtures of probabilistic principal component analyzers (MPPCA)~\citep{tipping_mixtures_1999} assume that each cluster lies around a low-dimensional affine subspace, following a multivariate Gaussian distribution with a specific eigenvalue profile---the smallest eigenvalues being \textit{constant} and equal to the noise level.
This model however overlooks an important difficulty which is the one of \textit{dimension selection}; the learning of the mixture parameters indeed requires to prespecify the intrinsic dimensions of the clusters, which are often unknown. The high-dimensional data clustering (HDDC)~\citep{bouveyron_high-dimensional_2007} and high-dimensional discriminant analysis (HDDA)~\citep{bouveyron_high-dimensional_2007-1} models address this issue by automatically adjusting the clusters dimensions over the iterations of the optimization algorithm, based on Cattell's scree test~\citep{cattell_scree_1966}. 
The theoretical guarantees of the learning algorithm are however not understood, and more practically, these models still require a hyperparameter as an input---the threshold for Cattell's scree test---which has an important impact on the choice of intrinsic dimensions. 

Other methods for the automatic learning of intrinsic dimensions in parsimonious GMMs exist. A natural approach adopted in~\cite{mcnicholas_parsimonious_2008} is the grid search combined with statistical model selection criteria such as the Akaike information criterion (AIC)~\citep{akaike_new_1974}, the Bayesian information criterion (BIC)~\citep{schwarz_estimating_1978}, or the integrated completed likelihood (ICL)~\citep{biernacki_assessing_2000}. While such a grid search is computationally expensive, few alternatives have been proposed: \citet{ghahramani_variational_1999} used a sparsity-inducing prior together with variational inference. In contrast, the problem of estimating the intrinsic dimension of a single Gaussian has received a much wider attention, using for instance likelihood-based techniques \citep{bouveyron_intrinsic_2011}, Bayesian model selection \citep{bishop_bayesian_1998,bouveyron_exact_2020}, or approximate cross-validation~\citep{josse_selecting_2012}.

Meanwhile, the building blocks of low-rank GMMs---the PPCA models of~\cite{tipping_probabilistic_1999}---have been extended into Gaussian models with piecewise-constant eigenvalue profiles under the name of \textit{principal subspace analysis} (PSA)~\citep{szwagier_curse_2025,szwagier_eigengap_2025}. The motivations are simple: if some sample eigenvalues are relatively close, then equalizing them slightly reduces the likelihood while \textit{drastically} reducing the number of parameters. In that sense, the PSA models go beyond low-rank models like PPCA in terms of parsimony and flexibility---since any group of eigenvalues can be equalized and not just the smallest ones. The eigenvalues multiplicities are automatically learned by optimizing a penalized log-likelihood criterion---e.g. the BIC. While PSA raises promising perspectives for parsimonious modeling and interpretability, it is certainly not suited for \textit{multimodal} density modeling---which is the case of most real datasets---or classical learning tasks such as classification and clustering.

In this paper, we address all the limits raised in the previous paragraphs at once with a new family of parsimonious GMMs termed \textit{mixtures of principal subspace analyzers} (MPSA). These models consist in the multimodal extension of PSA models, or equivalently the parsimonious generalization of MPPCA models with arbitrary eigenvalues multiplicities. At fixed multiplicities, we propose an expectation--maximization (EM) algorithm~\citep{dempster_maximum_1977} for the unsupervised learning of mixture parameters. 
When the multiplicities are not prespecified, we propose an efficient and provably-monotonous variant of the EM algorithm which automatically adjusts the groups of eigenvalues to equalize within the iterations. The underlying idea is that increasing the penalized log-likelihood \textit{componentwise} increases the \textit{mixture} penalized log-likelihood. Importantly, this algorithm is \textit{hyperparameter-free}, in the sense that the groups of eigenvalues to equalize do not depend on any hyperparameter to tune---compared to low-rank models like MPPCA and HDDC. As we shall discuss it, we believe that our EM variant can apply far beyond the scope of the current paper, providing new principles for integrated parameter estimation and model selection. 
We finally perform several experiments (density estimation, clustering, single-image denoising) which all show the superior likelihood--parsimony tradeoffs achieved by our models with respect to the classical GMMs.

\section{Principal subspace analysis}\label{sec:PSA}

PSA is a Gaussian model with a constraint on the covariance eigenvalues multiplicities.
It has been originally introduced to identify and overcome a ``curse of isotropy'' that arises in PCA when eigenvalues are relatively close.
In this section, we summarize these ideas through the lens of parsimonious Gaussian modeling. More details can be found in the original paper~\citep{szwagier_curse_2025}, including illustrations of the generative models, their poset structure, and the impact on the PCA methodology. 

Let 
\begin{equation}
    \mathcal{C}(p) \coloneqq \lrs{\gamma \in \mathbb{N}_*^d\colon \sum_{k=1}^d \gamma_k = p, \, d\in[\![1, p]\!]}    
\end{equation}
be the set of \emph{compositions} of $p \in \mathbb{N}_*$~\citep{bergeron_standard_1995}. Let $\gamma \in \mathcal{C}(p)$, $q_k \coloneqq \sum_{k'=1}^k \gamma_{k'}$, and let us write $q \coloneqq q_{d-1}$. 
The PSA model of \textit{type} $\gamma$ is defined as the parametric family of Gaussian densities $~{p(\,x \mid  \mu, \Sigma\,) = \mathcal{N}(\,x \mid  \mu, \Sigma\,)}$, where $\mu \in \R^p$ is the mean and $\Sigma \in \Sym_+(p)$ is a covariance matrix with repeated eigenvalues of multiplicities $\gamma$. 
More precisely, one assumes that the covariance matrix takes the form $~{\Sigma = \sum\nolimits_{k} \lambda_k \Pi_k}$, where $~{\lambda_k \in \R_+}$ is the $k$-th largest eigenvalue (of multiplicity $\gamma_k$) and $\Pi_k \in \Sym(p)$ is the orthogonal projector (of rank $\gamma_k$) onto the associated eigenspace (of dimension $\gamma_k$). Such a set of covariance matrices with eigenvalue multiplicities $\gamma$ is written $\Sym_+(\gamma)$. 

By definition, PSA models generalize the celebrated PPCA models of~\cite{tipping_probabilistic_1999}---whose $q$ highest eigenvalues are (almost surely) distinct and $p-q$ remaining eigenvalues are equal to the noise level---with types of the form $\gamma = (1, \dots, 1, p-q)$.\footnote{We will sometimes use the notation $1^a$, as an abbreviation for $1, \dots, 1$ (where 1 is repeated $a$ times).
}
PSA models also generalize the isotropic PPCA models of~\cite{bouveyron_intrinsic_2011}---whose eigenvalue profile forms two blocks, i.e. with types of the form $\gamma = (q, p-q)$.

PSA models are parameterized by the mean $~{\mu \in \R^p}$ and the covariance $\Sigma \in \Sym_+(\gamma)$. One could therefore think that the number of parameters is $~{p + {p(p+1)} / {2}}$. However, the repeated eigenvalues make the \textit{actual} number of parameters \textit{much smaller}. \cite{szwagier_curse_2025} show that the number of {free} parameters for a PSA model of type $\gamma$ is 
\begin{equation}\label{eq:PSA_kappa}
    \kappa(\gamma) = p + d + \frac{p^2 - \sum_{k=1}^d \gamma_k^2}{2}.
\end{equation}
Therefore, equalizing $\gamma_k$ eigenvalues decreases the number of parameters \textit{quadratically} ($O(\gamma_k^2)$).

Given a dataset $\lrp{x_i}_{i=1}^n$ in $\R^p$, the PSA model of type $\gamma$ that is the most likely to have generated the data (i.e. the \textit{maximum likelihood estimate}, MLE) can be computed in closed-form~\citep{szwagier_curse_2025}. The optimal mean is the empirical mean $\hat{\mu} = \frac1n \sum_{i=1}^n x_i$ and the optimal covariance matrix $\hat\Sigma$ is a piecewise-constant version of the empirical covariance matrix, where the eigenvalues have been \textit{block-averaged} according to $\gamma$. More precisely, if the empirical covariance $S \coloneqq \frac1n \sum_{i=1}^n (x_i - \bar{x})(x_i - \bar{x})\T$ eigendecomposes as $S = \sum_{j=1}^p \ell_j v_j {v_j}\T$, where $\ell_1 \geq \dots \geq \ell_p > 0$ are the eigenvalues\footnote{In the whole paper, for simplicity, we assume that the sample covariance matrices are always invertible. We make this assumption satisfied in practice by adding a \textit{covariance regularization} term to the diagonal of the covariance matrices.} and $v_1 \perp \dots \perp v_p$ are the eigenvectors, then
\begin{equation}\label{eq:PSA_ML}
    \hat{\Sigma}(\gamma) = \sum_{k=1}^d \hat\lambda_k(\gamma) \hat \Pi_k(\gamma),
\end{equation}
where 
\begin{align}
    \hat\lambda_k(\gamma) &= \sum_{j= q_{k-1} + 1}^{q_k} \frac{\ell_j}{\gamma_k},\\
    \hat \Pi_k(\gamma) &= \sum_{j= q_{k-1} + 1}^{q_k} v_j {v_j}\T
\end{align}
are respectively the block-averaged eigenvalues and their associated eigenspaces.
The maximum log-likelihood then writes
\begin{equation}\label{eq:PSA_LL}
	\ln\mathcal{L}(\hat\mu, \hat\Sigma(\gamma)) = -\frac{n}{2} \Biggl(p\ln n + \sum_{k=1}^d \gamma_k \ln\hat\lambda_k(\gamma) + p\Biggr).
\end{equation}

\noindent The BIC is defined as
\begin{equation}\label{eq:PSA_BIC}
	\operatorname{BIC}(\gamma) \coloneqq \kappa(\gamma) \ln n - 2 \ln\mathcal{L}(\hat\mu, \hat\Sigma(\gamma)).
\end{equation}
This model selection criterion results from an \textit{asymptotic} approximation of the model evidence and it makes a \textit{tradeoff} between \textit{model complexity} and \textit{goodness-of-fit}. 
Using the BIC, \cite{szwagier_curse_2025} get the important result that two adjacent sample eigenvalues $\ell_j$ and $\ell_{j+1}$ should be equalized when their \textit{relative eigengap} $\delta(\ell_j, \ell_{j+1}) \coloneqq \frac{\ell_j - \ell_{j+1}}{\ell_j}$ is below a certain threshold $\delta(n)$ depending on the number of samples $n$ but not on the dimension $p$:
\begin{equation}\label{eq:PSA_threshold}
	 \delta(n) \coloneqq 2 \lrp{1 - n^{2 / n} + n^{1 / n}\sqrt{n^{2 / n} - 1}}.
\end{equation}
Through extensive empirical experiments, \cite{szwagier_curse_2025} show that one rarely has enough samples to distinguish all the empirical eigenvalues, and that one should therefore transition from \textit{principal components} to \textit{principal subspaces}.

This important relative eigengap result suggests two simple methodologies to choose the groups of eigenvalues to equalize.
The \textit{relative eigengap strategy} consists in block-averaging the groups of eigenvalues whose relative eigengap is below the threshold $\delta(n)$.
The \textit{hierarchical clustering strategy} consists in a hierarchical clustering~\citep{ward_hierarchical_1963} of the eigenvalues, yielding a family of models $\mathcal{M}_\mathrm{h}(\ell)\in\mathcal{C}(p)^p$ of decreasing complexity, from which one chooses the model with the lowest BIC. The detailed algorithm is given in~\cite{szwagier_curse_2025}.
While the relative eigengap strategy is fast ($O(p)$) compared to the hierarchical clustering strategy ($O(p^2)$), one can show that the former always selects a PSA model with greater or equal BIC, since the latter optimizes over a larger set.\footnote{Indeed the (single-linkage) hierarchical clustering boils down to regrouping the eigenvalues sequentially according to their relative eigengap. Therefore, at some point, all the eigenvalue pairs below the relative eigengap threshold~\eqref{eq:PSA_threshold} are clustered, similarly to the relative eigengap strategy. In other words, the family $\mathcal{M}_\mathrm{h}(\ell)\in\mathcal{C}(p)^p$ contains the model output by the relative eigengap strategy.
}\label{foot:H_vs_R}
The more recent eigengap sparsity of~\cite{szwagier_eigengap_2025} relaxes the discrete BIC minimization problem into a continuous problem penalized by pairwise distances between sample eigenvalues~\citep{tyler_lassoing_2020}.

While PSA has an important impact on the \textit{interpretability} of principal components, its linear-Gaussian nature is too simplistic for real datasets. Moreover, it is not meant to be used for classical learning tasks such as classification and clustering.
Since PSA is nothing but a parsimonious Gaussian model, it could naturally be extended into a parsimonious Gaussian \textit{mixture} model.
This is the purpose of next section. As we shall see, such a seemingly natural extension is actually not-so-straightforward from a technical point of view since practical difficulties emerge from the high-dimensional nature of the problem.
 
\section{Mixtures of principal subspace analyzers}\label{sec:MPSA}

In this section, we introduce the MPSA models and propose an EM algorithm for learning the mixture parameters in the unsupervised context, with typical applications in density modeling and clustering. The proofs can be found in Appendix~\ref{app:proofs}.

\subsection{Mixture density and parameters}
\begin{definition}[MPSA density]
Let $C \in \mathbb{N}^*$ be a number of clusters. Let $\gamma \coloneqq \lrp{\gamma_1, \dots, \gamma_C}\in\mathcal{C}(p)^C$ be a sequence of PSA types---i.e. for $c \in [\![1, C]\!]$, $~{\gamma_c \coloneqq (\gamma_{c1}, \dots, \gamma_{c d_c})\in\mathbb{N}^{d_c}_*}$ with $~{\sum_{k=1}^{d_c} \gamma_{ck} = p}$ and $d_c\in[\![1, p]\!]$.
The MPSA model of types $\gamma$ is defined as the parametric family of probability densities
\begin{equation}\label{eq:MPSA}
p(\,x \mid  \theta\,) \coloneqq \sum_{c=1}^C \pi_c \, \mathcal{N}(\,x \mid  \mu_c, \Sigma_c\,),
\end{equation}
where $\pi_c\in\R_{+}$, $\mu_c\in\R^p$ and $\Sigma_c \in \Sym_+(\gamma_c)$. We denote the parameters $~{\theta \coloneqq (\pi_c,\mu_c, \Sigma_c)_{c=1}^C \in \Theta(\gamma)}$.
\end{definition}
\begin{remark}[Generalization of parsimonious GMMs]
As PSA extends and unifies PPCA with other parsimonious Gaussian models (like IPPCA, spherical and full-Gaussian models), MPSA extends and unifies MPPCA with other classical parsimonious GMMs belonging to~\cite{bouveyron_high-dimensional_2007} and~\cite{mcnicholas_parsimonious_2008}.
\end{remark}

\noindent Since no parameter is shared between the mixture components, the total number of parameters can be directly obtained from Equation~\eqref{eq:PSA_kappa}.
\begin{proposition}[MPSA complexity]\label{prop:MPSA_kappa}
The number of parameters for the MPSA model of types $\gamma$ is
\begin{equation}\label{eq:MPSA_kappa}
	\kappa(\gamma) = C - 1 + \sum_{c=1}^C \kappa(\gamma_c).    
\end{equation}
\end{proposition}

\subsection{Inference and EM algorithm}\label{subsec:EM}
Let us consider the problem of mixture parameter estimation under the MPSA model of types $\gamma$. In the unsupervised setting, the class labels are unknown. The classical EM algorithm for mixture parameters estimation then consists in alternating between the computation of the expected labels given the mixture parameters (E-step) and the maximum likelihood estimation of the mixture parameters given the expected labels (M-step). The computational interest lies in the explicit solutions to the M-step, which can be performed \textit{componentwise}. Moreover, the EM algorithm is guaranteed to increase the mixture log-likelihood $\ln \mathcal{L}(\theta) \coloneqq \sum_{i=1}^n \ln p(\,x_i \mid  \theta\,)$ at each step, notably by virtue of Gibbs' inequality.
We summarize our EM algorithm for MPSA inference in Algorithm~\ref{alg:EM} and give the details of each step in the remainder of this subsection. The posterior probabilities of class membership ($t_{ci} \coloneqq \mathbb{P}(\,y_i=c\mid x_i\,)$, where $y_i\in{[\![1, C]\!]}$ is a cluster indicator variable), which are explicited later, are gathered in a matrix $t\in\R^{C\times n}$ for conciseness in the pseudo-code.
\begin{algorithm}
\caption{EM algorithm for MPSA}\label{alg:EM}
\begin{algorithmic}
\Require $(x_i)_{i=1}^n\in\R^{p\times n}, C\in\mathbb{N_*}, \gamma\in\mathcal{C}(p)^C$
\Ensure $\hat \theta \in \Theta(\gamma)$ \Comment{mixture parameters}
\State $y \gets \operatorname{KMeans}(x, C) \in {[\![1, C]\!]}^n$ \Comment{K-Means init.}
\State $t \gets \lrp{\mathbf{1}_{y_i = c}}_{c, i}$ \Comment{posterior proba.}
\For{$s$ = 1, 2, \dots}
    \State $\hat \theta \gets \operatorname{M-step}(t, \gamma)$ \Comment{thm.~\ref{thm:M_step}}
    \State $t \gets \operatorname{E-step}(\hat \theta)$ \Comment{thm.~\ref{thm:E_step}}
\EndFor
\end{algorithmic}
\end{algorithm}
\begin{remark}[Supervised setting]
One can directly use Algorithm~\ref{alg:EM} for classification (supervised setting) by initializing the cluster indexes with the \textit{true} labels rather than with K-Means, and running only one iteration.
\end{remark}

\subsubsection{E-step}
Let us assume that a dataset $(x_i)_{i=1}^n$ has been sampled from an MPSA model of types $\gamma$ with known parameters $\theta$---i.e. according to the mixture density of Equation~\eqref{eq:MPSA}.
Then the probability that a data point $x\in\R^p$ belongs to cluster $c \in [\![1, C]\!]$ can be computed using Bayes' theorem:
\begin{equation}
    \mathbb{P}(\,y=c\mid x, \theta\,) = \frac{\pi_c \, \mathcal{N}(\,x\mid \mu_c, \Sigma_c\,)}{\sum_{c'=1}^C \pi_{c'} \, \mathcal{N}(\,x\mid \mu_{c'}, \Sigma_{c'}\,)}.
\end{equation}
We simplify this formula in the next theorem, referred to as \textit{E-step}, based on the eigendecomposition of the covariance matrices.
\begin{theorem}[E-step]\label{thm:E_step}
Consider an MPSA model of types $\gamma$ with parameters $\theta$. 
Let the covariance matrices decompose as $\Sigma_c = \sum_{k=1}^{d_c} \lambda_{ck} \Pi_{ck}$, where $\lambda_{ck}$ are the (ordered decreasing) eigenvalues and $\Pi_{ck}$ the associated eigenspaces, of dimension $\gamma_{ck}$.
Then the probability that $x\in\R^p$ belongs to cluster $c \in [\![1, C]\!]$ is
\begin{equation}\label{eq:E_step}
	\mathbb{P}(\,y=c\mid x, \theta\,) = \frac1{\sum_{c'=1}^C \exp\lrp{\frac{K_c(x) - K_{c'}(x)}2}},
\end{equation}
where $K_c$ corresponds to the cost function
\begin{multline}\label{eq:E_step_Kc}
K_c(x) = - 2\ln{\pi_c} + \sum_{k=1}^{d_c} \gamma_{ck}\ln{\lambda_{ck}}\\ 
+ \sum_{k=1}^{d_c} \frac{(x - \mu_c)\T \Pi_{ck} (x - \mu_c)}{\lambda_{ck}}.    
\end{multline}
The log-likelihood of a dataset $(x_i)_{i=1}^n$ then writes 
\begin{equation}\label{eq:LL}
	\ln\mathcal{L}(\theta) = \sum_{i=1}^n \ln\lrp{\sum_{c=1}^C \exp\lrp{-\frac{K_c(x_i) + p \ln 2\pi}{2}}}.
\end{equation}
\end{theorem}

\subsubsection{M-step}
Let us now assume that a dataset $(x_i)_{i=1}^n$ has been sampled from an MPSA model of types $\gamma$ with unknown parameters. 
In the supervised context, one has access to the class-labels $y_i \in [\![1, C]\!]$, so that the mixture parameters can be estimated by maximizing the \textit{complete-data} log-likelihood
\begin{equation}
    \ln\mathcal{L}_{\mathrm{c}}(\hat\theta) \coloneqq
    \sum_{c, i} \mathbf 1_{y_i=c} \lrp{\ln\hat\pi_c + \ln\mathcal N( \, x_i \mid \hat\mu_c, \hat\Sigma_c \,)}.
\end{equation}
In the unsupervised context, one does not have access to the labels, but the previous Theorem~\ref{thm:E_step} (E-step) gives a probabilistic estimate $~{t_{ci} \coloneqq \mathbb{P}(\,y_i=c\mid x_i, 
\theta\,)}$. 
Therefore, as classically done, we estimate the mixture parameters by maximizing the \textit{expected complete-data log-likelihood}
\begin{equation}\label{eq:expected_ll}
    \mathbb{E}[\ln\mathcal{L}_{\mathrm{c}}(\hat\theta)] = \sum_{c, i} t_{ci} \lrp{\ln\hat\pi_c + \ln\mathcal N( \, x_i \mid \hat\mu_c, \hat\Sigma_c \,)}.
\end{equation}
\begin{theorem}[M-step]\label{thm:M_step}
	Let $(x_i)_{i=1}^n$ be a dataset in $\R^p$, and $t_{ci} \coloneqq \mathbb{P}(\,y_i=c\mid x_i\,)$ the expected class labels.
	The MPSA parameters that maximize the expected complete-data log-likelihood~\eqref{eq:expected_ll} are
\begin{align}\label{eq:M_step}
    	\hat\pi_c(t_c) &= \frac1n \sum\limits_{i=1}^n t_{ci},\\
    	\hat\mu_c(t_c) &= \frac1{n \hat \pi_c} \sum\limits_{i=1}^n t_{ci} \, x_i,\\
    	\hat\Sigma_c(t_c, \gamma_c) &= \sum\limits_{k=1}^{d_c} \hat\lambda_{ck}(t_c, \gamma_c) \, \hat \Pi_{ck}(t_c, \gamma_c),
\end{align}
where 
$\hat\lambda_{ck}(t_c, \gamma_c)$ and $\hat \Pi_{ck}(t_c, \gamma_c)$
are the block-averaged eigenvalues and eigenspaces of the covariance matrix
\begin{equation}\label{eq:M_step_S}
	S_c (t_c) \coloneqq \frac1{n \hat \pi_c} \sum_{i=1}^n t_{ci} \lrp{x_i - \hat\mu_c} \lrp{x_i - \hat\mu_c}\T.
\end{equation}
More precisely, if $\ell_{c1} \geq \dots \geq \ell_{cp}$ are the eigenvalues of $S_c$ and $v_{c1} \perp \dots \perp v_{cp}$ are the associated eigenvectors, then one has 
\begin{align}
    \hat\lambda_{ck}(t_c, \gamma_c) &= \sum_{j= q_{c(k-1)} + 1}^{q_{ck}} \frac{\ell_{cj}(t_c)}{\gamma_{ck}},\\
    \hat \Pi_{ck}(t_c, \gamma_c) &= \sum_{j= q_{c(k-1)} + 1}^{q_{ck}} v_{cj}(t_c) {v_{cj}(t_c)}\T.
\end{align}
\end{theorem}
\noindent Hence, the M-step consists in a classical Gaussian mixture M-step (i.e. computing the local responsibility-weighted covariance matrices $S_c$) followed by a PSA inference step (i.e. block-averaging the eigenvalues according the pre-specified eigenvalues multiplicities $\gamma_c$).

\subsection{EM complexity analysis}~\label{subsec:complexity}
We now analyze the time complexity of the proposed EM algorithm~\ref{alg:EM}. For simplicity, we assume that $n > p$. The case $n \leq p$ can be tackled via transpose tricks combined with covariance regularization to avoid singular covariance matrices.

The K-Means clustering algorithm~\citep{lloyd_least_1982} involved in the EM initialization has time complexity $O(C n p T)$, with $T$ the number of iterations, which is by default at most $300$ in the scikit-learn implementation~\citep{pedregosa_scikit-learn_2011}.

A classical E-step would have time complexity $O(C p^3)$ due to the covariance inversion.
However, the formulas that we derive in Theorem~\ref{thm:E_step} have several computational advantages.
First, they remove the necessity to numerically invert the covariance matrices---which can be pathological in high dimensions. Since the eigenvalues $\lambda_{ck}\in\R$ and the eigenvectors $~{Q_{ck}\in\R^{p\times \gamma_{ck}}}$ are computed in the M-step, the computational complexity of the cost $K_c(x)$ is the one of $~{\sum_k {\lambda_{ck}}^{-1}(x - \mu_c)\T \Pi_{ck} (x - \mu_c)}$, which can be rewritten as $~{\sum_{k} {\lambda_{ck}}^{-1} \|{Q_{ck}}\T(x-\mu_c)\|_2^2}$, that is $\sum_{k=1}^{d_c} O(\gamma_{ck}p) = O(p^2)$. Therefore, the complexity of the E-step is $O(C n p^2)$. But this is not all.
Since $(\Pi_{c1}, \dots, \Pi_{cd_c})$ represents a \textit{flag} of subspaces~\citep{ye_optimization_2022}---i.e. mutually-orthogonal subspaces spanning the whole space $\R^p$---then any orthogonal projection matrix $\Pi_{ck}$ can be expressed as $\Pi_{ck} = I - \sum_{k'\neq k} \Pi_{ck'}$. Notably, if one projector is of much greater rank than the other ones (for instance the last one $\Pi_{c d_c}$, accounting for the noise), then it can be replaced with lower dimensional terms in~\eqref{eq:E_step_Kc} and yield much more computationally-efficient formulas for the E-step.
More precisely, if one has $~{q_c \coloneqq p - \gamma_{c d_c} = O(1)}$, then the computational complexity of the cost function $K_c$ drops to $O(q_c p) = O(p)$.
Eventually, the complexity of the E-step in high dimensions will more likely be $O(C n p q)$, with $~{q \coloneqq \max_{c\in[\![1, C]\!]} q_c}$.

The M-step involves the eigenvalue decomposition (EVD) of the local responsibility-weighted covariance matrices $S_c$. As currently implemented for simplicity, computing both the covariance matrices and their EVDs results in a complexity of $O(Cnp^2) + O(Cp^3)$. Therefore, even though the eigenvalue equalization of Theorem~\ref{thm:M_step} is extremely simple, it does not remove the need for the costly EVD. To make it more efficient in high dimensions, one can use classical numerical tricks such as the truncated singular value decomposition (SVD), which computes only a few leading eigenvectors. This perspective is investigated in Section~\ref{subsec:TSVD}.

In Algorithm~\ref{alg:EM}, the mixture model types $\gamma$ are assumed to be prespecified. Unless they are known, a classical way to choose these hyperparameters is via cross-validation or statistical model selection techniques, implying running several EM algorithms~\citep{mcnicholas_parsimonious_2008}.
The important issue here is that the number of candidate MPSA models is ${(2^{p-1})}^C$.
Indeed, each independent mixture component is parameterized by its type $\gamma_c$ which can take $2^{p-1}$ values, since each adjacent-eigenvalue pair can be either equalized or not. Therefore, an exhaustive model selection would imply running $O(2^{Cp})$ EM algorithms, which becomes computationally prohibitive for $p \gtrapprox 10$---even though the EM itself may not be too much computationally costly after integrating the previously-exposed low-dimensional numerical tricks.

Drawing from the HDDC models of~\cite{bouveyron_high-dimensional_2007}, one could naturally be tempted to perform some kind of model selection at the component level (i.e. based on the PSA methodology of~\cite{szwagier_curse_2025} presented in Section~\ref{sec:PSA}) at each EM iteration. This would add a negligible time complexity to Algorithm~\ref{alg:EM} ($\O(C p^2)$ for the hierarchical clustering strategy), instead of a multiplicative factor.
However, such a variant of the EM algorithm would not enjoy any theoretical guarantee a priori. 
Consequently, we propose in the next section to reframe our intuitions as a penalized optimization problem, for which we show that a thoughtful variant of the EM algorithm yields a monotonous loss function.

\section{A penalized EM algorithm}\label{sec:CPEM}
We propose to reframe the joint learning of the mixture parameters and hyperparameters (eigenvalue multiplicities) as a \textit{penalized log-likelihood} optimization problem:
\begin{multline}\label{eq:pen_lik}
    \argmax_{\substack{\gamma\in\mathcal{C}(p)^C\\ (\pi_c,\mu_c,\Sigma_c)_{c=1}^C \in \Theta(\gamma)}}
    \sum_{i=1}^n \ln \lrp{\sum_{c=1}^C \pi_c \, \mathcal{N}(\,x_i \mid  \mu_c, \Sigma_c\,)}\\
    - \alpha \, \kappa(\gamma),
\end{multline}
where $\alpha\in\R_+$ is a regularization hyperparameter (for instance $\alpha = \ln(n)/2$ for a BIC-like penalized log-likelihood).

\subsection{Componentwise monotonicity implies mixture monotonicity}
It is well known from the classical EM theory~\citep{dempster_maximum_1977} that increasing the expected complete-data log-likelihood $\mathbb{E}_{\,y\mid x,\theta\,} [\ln\mathcal{L}_{\mathrm{c}}(\cdot)]$~\eqref{eq:expected_ll} increases the mixture log-likelihood $\ln\mathcal{L}(\cdot)$~\eqref{eq:LL}. This is notably due to Gibbs' inequality---i.e. the non-negativity of the relative entropy. In this work, we are not interested in increasing the log-likelihood but a penalized version of it~\eqref{eq:pen_lik}, to enforce parsimonious modeling. As said earlier, we would like to select the eigenvalues multiplicities with componentwise model selection strategies drawn from the PSA methodology. However, such a variant of the EM algorithm would not enjoy a priori any guarantee on the evolution of the objective function~\eqref{eq:pen_lik} over the iterations.
The following theorem states that a thoroughly-designed variant of the EM algorithm where we replace the M-step with a componentwise model selection guarantees the monotonicity of the penalized log-likelihood.
We call this variant \textit{componentwise penalized EM} (CPEM).
\begin{theorem}[CPEM]\label{thm:CPEM}
    Let $\gamma\in\mathcal{C}(p)^C$, $\theta\in\Theta(\gamma)$, $~{t_{ci} \coloneqq \mathbb{P}(\,y_i=c\mid x_i, \theta\,)}$ (cf.~Thm.~\ref{thm:E_step}) and
    $\Psi_c(\gamma_c) \coloneqq \sum_{i=1}^n t_{ci} \ln\mathcal N( \, x_i \mid \hat{\mu}_c(t_c), \hat\Sigma_c(t_c, \gamma_c)\,)$ (cf.~Thm.~\ref{thm:M_step}).
    Let $\gamma'\in\mathcal{C}(p)^C$ such that $\forall c \in [\![1, C]\!]$,
    \begin{equation}\label{eq:component_increase}
        \Psi_c(\gamma_c') -  \alpha \, \kappa(\gamma_c') \geq \Psi_c(\gamma_c) - \alpha \, \kappa(\gamma_c),
    \end{equation}
    and let $\theta(\gamma') \coloneqq (\hat\pi_c(t_c), \hat\mu_c(t_c), \hat\Sigma_c(t_c, \gamma'_c))_{c \in [\![1, C]\!]}$. Then one has:
    \begin{equation}
        \ln \mathcal{L}(\theta(\gamma')) - \alpha \, \kappa(\gamma') \geq
        \ln \mathcal{L}(\theta) - \alpha \, \kappa(\gamma).
    \end{equation}
\end{theorem}
\noindent In other words, thoroughly updating the eigenvalue multiplicities componentwise via~\eqref{eq:component_increase} guarantees the monotonicity of the objective function~\eqref{eq:pen_lik} over the iterations of the EM algorithm.
\begin{remark}[Simplification of componentwise inequalities]
    Using Theorem~\ref{thm:E_step}, one can show with a similar reasoning to Equation~\eqref{eq:PSA_LL} that $~{\Psi_c(\gamma_c) = -\frac{n\pi_c}{2} \sum_{k=1}^{d_c} \gamma_{ck} \ln\hat\lambda_{ck}(t_c, \gamma_c) + \mathrm{cst}}$, which simplifies the implementation.
    For instance, with a BIC-like penalization ($\alpha = \ln(n)/2$), the problem of model selection~\eqref{eq:component_increase} can simply be formulated as 
    \begin{equation}\label{eq:CMS}
        \argmin_{\gamma_c\in\mathcal{C}(p)} \, \sum_{k=1}^{d_c} \gamma_{ck} \ln\hat\lambda_{ck}(t_c, \gamma_c) + \kappa(\gamma_c) \frac{\ln n}{n\pi_c}.
    \end{equation}
\end{remark}
\begin{remark}[Impact on low-rank mixture models]
    Theorem~\ref{thm:CPEM} applies straightforwardly to other parsimonious GMMs such as the mixtures of PPCA of~\cite{tipping_mixtures_1999}, the HDDC models of~\cite{bouveyron_high-dimensional_2007} and the parsimonious models of~\cite{mcnicholas_parsimonious_2008} (which the MPSA models generalize), by choosing types of the form $\gamma_c = (1, \dots, 1, p - q_c)$. Therefore, our results implicitly address the important issue of automatic intrinsic dimension selection in low-rank mixture models that we raised in Section~\ref{sec:intro}. 
    At each step of the EM, for each mixture component $c$, we could select the intrinsic dimension $q_c$ that solves 
    \begin{multline}\label{eq:CMS_PPCA}
        \argmin_{q\in\{0, \dots, p-1\}} \, \sum_{j=1}^{q}  \ln\ell_{cj} + (p - q) \ln\Biggl(\frac{1}{p-q}\sum_{j=q+1}^{p} \ell_{cj}\Biggr) \\
        + \lrp{pq-\frac{q(q-1)}{2}} \frac{\ln n}{n\pi_c}. 
    \end{multline}
    The point is that our MPSA models \textit{contain} the MPPCA models, so that solving~\eqref{eq:CMS}---which is what we do in this paper---always yields more adapted models than what we could get from solving~\eqref{eq:CMS_PPCA}.
\end{remark}
\begin{remark}[Novelty of CPEM]
As far as we know, our componentwise penalized EM monotonicity theorem is new. 
We believe that it can have an important impact in the parsimonious mixture models community~\citep{tipping_mixtures_1999,mcnicholas_parsimonious_2008,scrucca_mclust_2016}. Indeed, the problem of hyperparameter selection in mixture models is notoriously challenging and open, and it is often addressed heuristically without theoretical guarantees~\citep{bouveyron_high-dimensional_2007}. Therefore, the monotonicity result that we obtain addresses this important issue.
Let us note that the idea of integrating both parameter estimation and model selection in an EM algorithm is absolutely not new; it was notably proposed in the seminal work of~\cite{figueiredo_unsupervised_2002} for finite mixture models and theoretically investigated in~\cite{jiang_e-ms_2015} for missing data.
Let us also note that our algorithm has little to do with the componentwise EM of~\cite{celeux_componentwise_2001}, which updates the mixture parameters sequentially, one component per iteration.
\end{remark}

\subsection{Algorithm}\label{subsec:CPEM_algo}
Theorem~\ref{thm:CPEM} motivates us to design a variant of the EM algorithm where the componentwise quantities $\Psi_c(\cdot) -  \alpha \, \kappa(\cdot)$ increase at each EM iteration.
For that, at each step $s$, for each mixture component $c$, we consider a set of candidate models $\mathcal{M}(\gamma_c)\subset\mathcal{C}(p)$ and take the one that maximizes $\Psi_c(\cdot) -  \alpha \, \kappa(\cdot)$. The general methodology is summarized in Algorithm~\ref{alg:CPEM}.
\begin{algorithm}
\caption{CPEM algorithm for MPSA}\label{alg:CPEM}
\begin{algorithmic}
\Require $(x_i)_{i=1}^n\in\R^{p\times n}, C\in\mathbb{N_*}, \gamma_0\in\mathcal{C}(p)^C$
\Ensure $\hat \gamma \in \mathcal{C}(p)^C, \hat \theta \in \Theta(\hat \gamma)$
\State $y \gets \operatorname{KMeans}(x, C) \in {[\![1, C]\!]}^n$ \Comment{K-Means init.}
\State $t \gets \lrp{\mathbf{1}_{y_i = c}}_{c, i}$ \Comment{posterior proba.} 
\State $\hat\gamma \gets \gamma_0$ \Comment{types init. (spherical or full)}
\For{$s$ = 1, 2, \dots}
    \State $\hat\gamma \gets \argmax\limits_{\gamma_c\in \{\hat\gamma_c\} \cup \mathcal{M}(\hat\gamma_c)} \Psi_c(\gamma_c) -  \alpha \, \kappa(\gamma_c)$ \Comment{thm.~\ref{thm:CPEM}}
    \State $\hat \theta \gets \operatorname{M-step}(t, \hat\gamma)$ \Comment{thm.~\ref{thm:M_step}}
    \State $t \gets \operatorname{E-step}(\hat \theta)$ \Comment{thm.~\ref{thm:E_step}}
\EndFor
\end{algorithmic}
\end{algorithm}
\begin{remark}[Satisfying monotonicity conditions]\label{rem:monotonicity}
In order to ensure that the theoretical monotonicity conditions from Theorem~\ref{thm:CPEM} are met at each iteration---regardless of the set of candidate models $\mathcal{M}(\hat\gamma_c)$---we include the current model $\hat\gamma_c$ in the set of candidate models. Indeed, not switching model at a CPEM iteration corresponds to a classical M-step from the EM algorithm and it naturally satisfies Equation~\eqref{eq:component_increase} (with $\gamma'_c = \gamma_c$). Consequently, the selected model, $\argmax_{\gamma_c\in \{\hat\gamma_c\} \cup \mathcal{M}(\hat\gamma_c)} \Psi_c(\gamma_c) -  \alpha \, \kappa(\gamma_c)$ satisfies the theoretical monotonicity conditions too, a fortiori.
\end{remark}

\noindent There are many possibilities for the choice of candidate models $\mathcal{M}(\gamma_c)$---the idea being of course to avoid considering all the possible models, i.e. $\mathcal{M}(\gamma_c) = \mathcal{C}(p)$, in which case we would have $2^{p-1}$ models to compare at each step $C$. Two ideas have been mentioned in Section~\ref{sec:PSA}: the \textit{relative eigengap strategy}---for which $\mathcal{M}(\gamma_c)$ consists in the singleton containing the model obtained by clustering the eigenvalues  of $S_c$ whose relative distance is below the threshold~\eqref{eq:PSA_threshold}---and the \textit{hierarchical clustering strategy}---for which $\mathcal{M}(\gamma_c)$ consists in the union of models obtained by the hierarchical clustering of the eigenvalues $\mathcal{M}_\mathrm{h}(\ell_c)\in\mathcal{C}(p)^p$, the metric being the relative distance.

We now introduce two additional ideas, borrowing from classical paradigms of model selection and sometimes referred to as \textit{top-down} and \textit{bottom-up} approaches (see \cite{bach_optimization_2012} for uses in sparse optimization problems and \cite{verbockhaven_growing_2024} for uses in {neural architecture search}). 
Let $\gamma \coloneqq (\gamma_1, \dots, \gamma_d) \in \mathcal{C}(p)$. We say that $\gamma'\in\mathcal{C}(p)$ is an \textit{upper neighbor} of $\gamma$ and write $\gamma'\in\mathcal{N}^+(\gamma)$ if it can be expressed as $\gamma' = (\gamma_1, \dots, \gamma_{k-1}, \gamma_{k1}, \gamma_{k2}, \gamma_{k+1}, \dots, \gamma_d)$ with $(\gamma_{k1}, \gamma_{k2})\in\mathcal{C}(\gamma_k)$, for any $k\in\{1, \dots, d\}$. Similarly, we say that $\gamma'\in\mathcal{C}(p)$ is a \textit{lower neighbor} of $\gamma$ and write $\gamma'\in\mathcal{N}^-(\gamma)$ if it can be expressed as $\gamma' = (\gamma_1, \dots, \gamma_{k-1}, \gamma_k + \gamma_{k+1}, \gamma_{k+2}, \dots, \gamma_d)$. One can easily show that $\#\mathcal{N}^+(\gamma) = p - d$ and $\#\mathcal{N}^-(\gamma) = d - 1$.\footnote{
These definitions are connected to the notion of \textit{type refinement} introduced in~\cite{szwagier_curse_2025}, and they can be more-intuitively visualized via the Hasse diagram representation of PSA models from the paper.}
The \textit{bottom-up strategy} consists in starting from spherical GMMs ($\gamma_c = (p) \, \forall c$) and increasing step by step the model complexity by finding the ``split'' in the eigenvalue profile that maximizes the penalized log-likelihood (i.e. taking $\mathcal{M}(\gamma_c) = \mathcal{N}^+(\gamma_c)$). The \textit{top-down strategy} is similar but instead starts from the full GMMs ($\gamma_c = (1, \dots, 1) \, \forall c$) and decreases the complexity at each step by equalizing the two adjacent groups of eigenvalues that maximize the penalized log-likelihood (i.e. taking $\mathcal{M}(\gamma_c) = \mathcal{N}^{-}(\gamma_c)$). 
\begin{remark}[Up and down]\label{rem:up_down}
    Early experiments showed that the above-defined bottom-up and top-down strategies may get stuck in local optima due to their monotonous behavior in terms of model complexity. For instance, the bottom-up approach may increase too much the number of parameters during the first few iterations (similarly to~\cite{houdard_high-dimensional_2018}, Figure~7) because one needs a lot of degrees of freedom to model the density when the mixture centers are initialized far from the true cluster centers. Once the true centers are found, one may realize that the clusters have a much simpler structure and one would like to reduce the number of parameters (as we can see on the bottom of Figure~\ref{fig:clus_strategies} for the red curve). 
    To enable such late changes, we eventually take $\mathcal{M}(\gamma_c) = \mathcal{N}^+(\gamma_c) \cup \mathcal{N}^-(\gamma_c)$ for both bottom-up and top-down strategies.
\end{remark}

\begin{remark}[Which heuristic to choose?]
    Four strategies for updating the eigenvalue multiplicities during the CPEM algorithm have been proposed in this subsection, all of which satisfy the theoretical monotonicity conditions (cf. Remark~\ref{rem:monotonicity}). While theory shows that each strategy yields a non-decreasing sequence of penalized log-likelihoods, it does not show \textit{how much} the final parameters increase the penalized log-likelihood---with respect to initialization, to more classical GMMs, or in between the different strategies. Gibbs' inequality only implies that the penalized log-likelihood increases at least as much as its expected complete-data version. Therefore, one can expect complex strategies to outperform simple ones (for instance, the hierarchical clustering strategy should outperform the relative eigengap strategy, that it encompasses, as discussed in the end of Section~\ref{sec:PSA}) but one cannot really prove it. The empirical experiments of the next section aim to better understand the respective advantages of each strategy.
\end{remark}

\section{Experiments}
In this section, we perform various density modeling and clustering experiments with our new MPSA models (MPSA-H for the hierarchical clustering strategy, MPSA-R for the relative eigengap strategy, MPSA-U for the bottom-up strategy and MPSA-D for the top-down strategy), and compare them to the more standard full Gaussian mixture model (GMM-F) and the spherical one (GMM-S). The code is publicly available on GitHub: \url{https://github.com/tomszwagier/MPSA}. Technical implementation details can be found in the appendix (sections~\ref{subsec:TSVD} and~\ref{subsec:details}).

We notably perform some experiments on a synthetic dataset following the MPSA density $p(\, x \mid  \theta\,) = \sum_{c=1}^C \pi_c \, \mathcal{N}(\, x \mid  \mu_c, \Sigma_c \,)$~\eqref{eq:MPSA}. While $C$ and $\pi_c$ are arbitrary and specified for each experiment, the choice of other hyperparameters follow certain rules that we describe in this paragraph. The mean vectors are sampled uniformly between some prespecified bounds, i.e. $\mu_c\sim\mathcal{U}\lrp{\lrb{-b, b}^p}$. The covariance matrices $\Sigma_c = Q_c \Lambda_c {Q_c}\T$ are such that $Q_c$ is sampled uniformly (according to Haar's measure) in $\O(p)$, and $\Lambda_c$ has a constant relative eigengap $\delta$ between groups of eigenvalues, i.e. $\lrp{\lambda_{cj} - \lambda_{c (j+1)}} / {\lambda_{cj}} = \delta$ if $j=q_{ck}$, else $\lambda_{j} = \lambda_{j+1}$. The latter implying that $\lambda_{cp} / \lambda_{c1} = (1 - \delta)^{d_c-1}$, one then only needs to specify the highest eigenvalue $\lambda_{c1}$, the signal-to-noise ratio $\lambda_{cp} / \lambda_{c1}$ and the type $\gamma_c$ to fully determine the law of the covariance matrix $S_c$. We can now dive into the experiments.

\begin{remark}[Choice of $\alpha$]
    The penalty factor is set to $\alpha=\ln(n)/2$ in all the experiments, which corresponds to a BIC-like penalized log-likelihood. The choice of $\alpha$ certainly has an influence on the results---we can naturally expect the number of parameters of the selected model to decrease with $\alpha$. We conduct a sensitivity analysis in Section~\ref{subsec:alpha}, showing notably that one recovers the full GMM in the small $\alpha$ limit and the spherical GMM in the large $\alpha$ limit, and that the BIC tends to select (slightly) underparameterized models while the AIC tends to select (slightly) overparameterized models.
\end{remark}

\begin{remark}[Choice of $C$]
    The number of clusters is set to the true one (if known) in the following experiments. Indeed, this paper focuses on the parsimony induced by equalizing close covariance eigenvalues rather than reducing the number of mixture components. However, for completeness, we discuss in Section~\ref{subsec:C} how $C$ can be inferred asymptotically via a model selection framework or alternative methodologies relying on Dirichlet-type priors on the mixture proportions.
\end{remark}

\subsection{CPEM in action for 2D data}
\begin{figure}
    \centering
    \includegraphics[width=\linewidth]{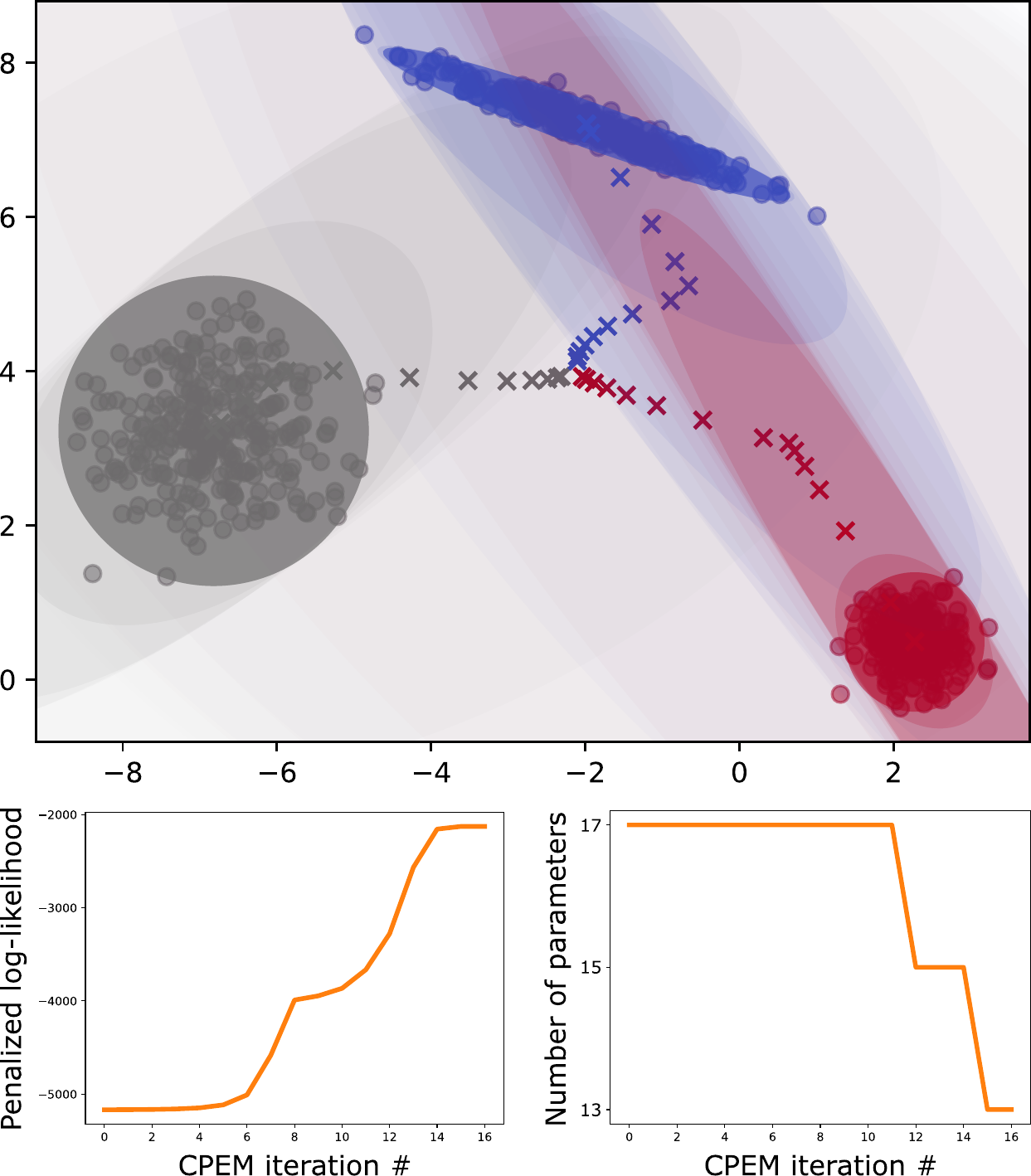}
    \caption{Two-dimensional density fitting with the MPSA-H. 
    Top: evolution of the mixture parameters over the CPEM iterations. The dots represent the data points, the crosses represent the evolution of the centers $\mu_c$ and the ellipses (with increasing opacity) represent the evolution of the covariance matrices.
    Bottom: evolution of the penalized log-likelihood (left) and the number of parameters (right) over the CPEM iterations.
    }
    \label{fig:2D_clus_1000}
\end{figure}

This experiment aims at understanding how the mixture parameters as well as the types evolve during the learning (Algorithm~\ref{alg:CPEM}) with a simple two-dimensional example. For that, we take $n=1000$, $p=2$, $C=3$, ${(\gamma_1, \gamma_2, \gamma_3) = ((1, 1), (2), (2))}$ (i.e. one anisotropic and two isotropic Gaussians), $(\pi_1, \pi_2, \pi_3) = (0.4, 0.3, 0.3)$, $\mu_c \sim \mathcal{U}\lrp{\lrb{-8, 8}^p}$, $(\lambda_{11}, \lambda_{21}, \lambda_{31}) = (1, .5, .1)$, and $\lambda_{cp} / \lambda_{c1} = 0.01 \, \forall c$, and we fit an MPSA model with the hierarchical clustering strategy.
We display in Figure~\ref{fig:2D_clus_1000} the scatter plot of the data and the learned mixture model, as well as the evolution of the penalized log-likelihood and the number of parameters.

We see that the clusters are well estimated, as well as their isotropic nature (for the gray and red ones) due to an appropriate MPSA type change occurring after a dozen iterations.
Moreover, the penalized log-likelihood increases at each iteration as expected by theory (cf. Theorem~\ref{thm:CPEM}). The number of parameters decreases over the iterations and ends up being the right one.\footnote{$~{\kappa((1, 1), (2), (2)) = 2 + 5 + 3 + 3 = 13}$ (cf. Proposition~\ref{prop:MPSA_kappa}).}

\subsection{Model selection strategies}
\begin{figure}
    \centering
    \includegraphics[width=\linewidth]{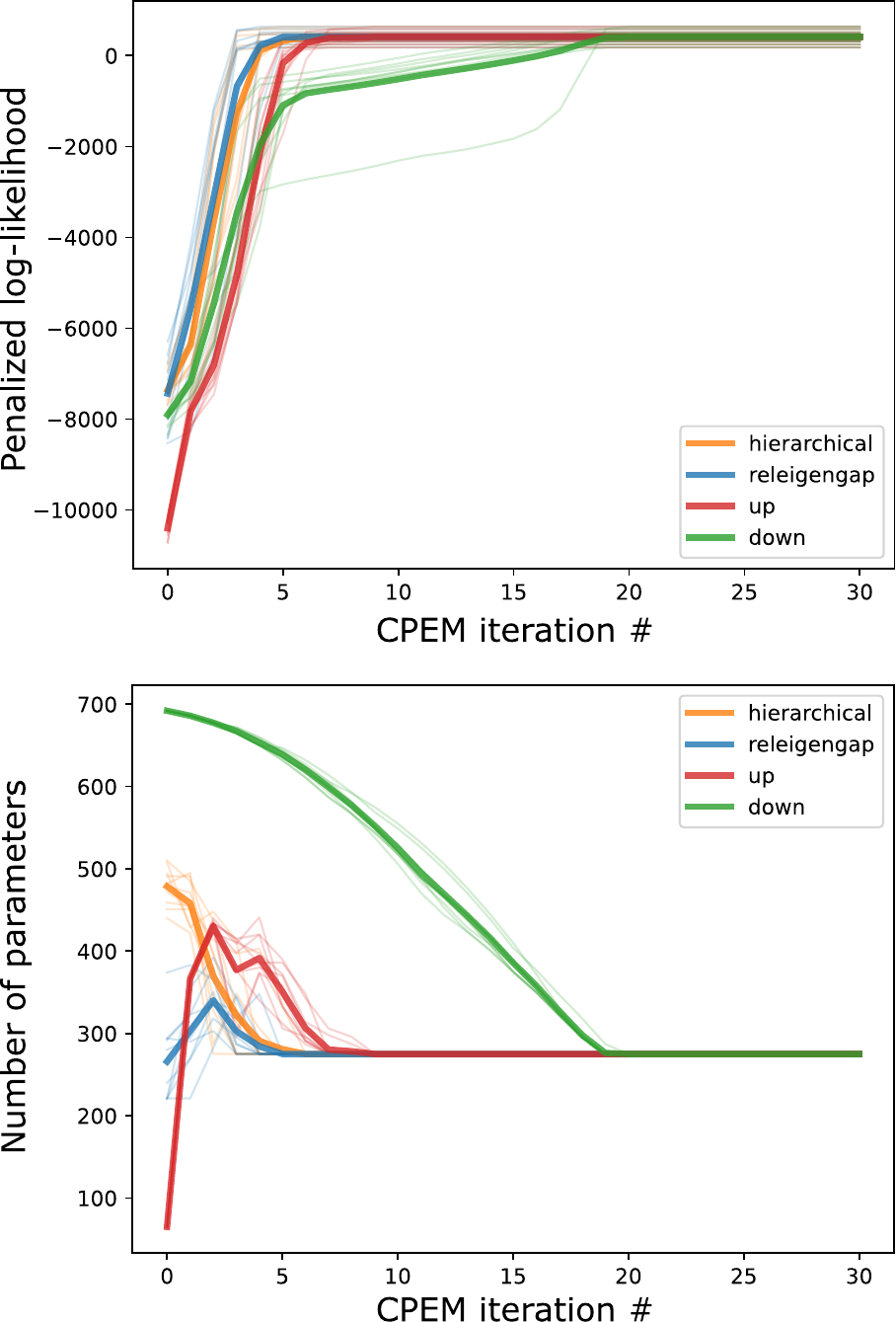}
    \caption{Evolution of the penalized log-likelihood and the number of parameters over the iterations of the CPEM algorithm for the four model selection strategies (MPSA-H/R/U/D).
    Each transparent curve refers to one repetition while the thick opaque curve represents the average over the 10 independent repetitions.
    }
    \label{fig:clus_strategies}
\end{figure}

The aim of this experiment is to study the learning algorithm~\ref{alg:CPEM} for different model selection strategies (MPSA-H/R/U/D) in higher dimension. To that extent, we plot in Figure~\ref{fig:clus_strategies} the penalized log-likelihood curves and the evolution of the number of parameters over the CPEM iterations, on a synthetic dataset with $n=1000$, $p=20$, $C=3$, ${(\gamma_1, \gamma_2, \gamma_3) = ((3, 5, 12), (8, 12), (20,))}$, $(\pi_1, \pi_2, \pi_3) = (0.4, 0.3, 0.3)$, $\mu_c = 0$, $(\lambda_{11}, \lambda_{21}, \lambda_{31}) = (1, .5, .1)$, and $\lambda_{cp} / \lambda_{c1} = 0.01 \, \forall c$, with 10 independent repetitions.
We see that the penalized log-likelihoods are once again monotonous, as expected by theory (cf. Theorem~\ref{thm:CPEM}). 

Let us now comment the results based on our intuitions.
All penalized log-likelihood curves seem to converge to the same optimum, after approximately 5 to 20 iterations. MPSA-D takes a much larger number of iterations to converge than the other strategies. However, the learning curves (penalized log-likelihood and number of parameters) are quite regular and with small variance, compared to the other strategies. We guess that in higher dimensions, MPSA-D may badly behave at first iterations due to the initialization with full covariance matrices, and it can moreover require a large number of iterations to converge, since it can only decrease the model complexity one by one. Therefore, smarter initialization heuristics could be designed, but this is left to future research.
MPSA-U also yields quite regular learning curves, although we can observe small oscillations of the number of parameters near the optimum (cf. Remark~\ref{rem:up_down} in Section~\ref{subsec:CPEM_algo}). 
We moreover guess that MPSA-U may get stuck in local optima due to the simplicity of the models at first iterations~\citep{verbockhaven_growing_2024}.
MPSA-H and MPSA-R are more flexible in terms of model complexity---which may make them attain the optimum faster as we can see in Figure~\ref{fig:clus_strategies}. However, we guess that their flexibility may also cause higher variance in performance and parsimony.

To deepen our understanding of the different strategies and compare them to the more standard GMMs, let us run some density modeling and clustering experiments at varying dimensions and distributions.

\subsection{Penalized log-likelihood benchmark}
The goal of this experiment is to compare our new parsimonious GMMs (MPSA-H/R/U/D) to the more-standard GMMs (full and spherical) in terms of penalized log-likelihood~\eqref{eq:pen_lik} $\sum_{i=1}^n \ln ({\sum_{c=1}^C \hat\pi_c \, \mathcal{N}(\,x_i \mid  \hat\mu_c, \hat\Sigma_c\,)}) - \alpha \, \kappa(\hat\gamma)$ (with $\alpha = \ln(n)/2$) under various settings. All datasets but the last one are synthetic, sampled according to the generative model presented in the beginning of this section. The hyperparameters are
$C=3$, $n=1000$, $(\pi_1, \pi_2, \pi_3) = (0.4, 0.3, 0.3)$, $\mu_c \sim \mathcal{U}\lrp{\lrb{-5, 5}^p}$, $(\lambda_{11}, \lambda_{21}, \lambda_{31}) = (3, 2, 1)$, $\lambda_{cp} / \lambda_{c1} = 0.01 \, \forall c$, and $p\in\{10, 100\}$.
For the first two datasets (MPSA $10$ and MPSA $100$), one takes ${\gamma_1, \gamma_2, \gamma_3 = (1, p-1), (1, 2, p-3), (1, 2, 4, p-7)}$ for $p\in\{10, 100\}$.
For the third and fourth dataset (Full $10$ and Full $100$), one takes ${(\gamma_1, \gamma_2, \gamma_3) = ((1^p), (1^p), (1^p))}$ for $p\in\{10, 100\}$, i.e. full Gaussian models with distinct population eigenvalues.
The Skew $100$ dataset has ${(\gamma_1, \gamma_2, \gamma_3) = ((1^p), (1^p), (1^p))}$ for $p=100$ and a skew-normal distribution (instead of the Gaussian one used in the previous experiments).
The last dataset (Ionosphere) is real~\citep{sigillito_ionosphere_1989}, with $C=2$, $n=351$ and $p=34$.
We report in Table~\ref{tab:clustering_penlik} the average penalized log-likelihood reached at the end of the CPEM algorithm, over 10 independent $n$-samples of the synthetic datasets. The last dataset---which is real---is split using a stratified $10$-fold cross-validation.
\begin{table*}[t]
\caption{Comparison of the penalized log-likelihoods $\sum_{i=1}^n \ln ({\sum_{c=1}^C \hat\pi_c \, \mathcal{N}(\,x_i \mid  \hat\mu_c, \hat\Sigma_c\,)}) - \alpha \, \kappa(\hat\gamma)$ reached by the MPSA-H/R/U/D models at the end of the CPEM algorithm, and the standard GMMs under various data distributions. The leading penalized log-likelihoods are in bold.
}
\label{tab:clustering_penlik}
\begin{footnotesize}
\begin{tabular*}{\textwidth}{@{\extracolsep\fill}lcccccc}
\toprule%
Model & MPSA $10$ & MPSA $100$ & Full $10$ & Full $100$ & Skew $100$ & Ionosphere\\
\midrule
MPSA-H & $\mathbf{-0.65 \pm 0.06}$ & $\mathbf{46.00 \pm 0.22}$ & $\mathbf{-07.59 \pm 0.06}$  & $\mathbf{-093.19 \pm 0.42}$ & $\mathbf{-093.98 \pm 2.82}$ & $\mathbf{+06.17 \pm 1.69}$\\
MPSA-R & $\mathbf{-0.65 \pm 0.06}$ & $\mathbf{46.00 \pm 0.22}$ & $\mathbf{-07.59 \pm 0.06}$  & $-100.02 \pm 0.28$ & $-099.09 \pm 0.72$ & $\mathbf{+04.76 \pm 1.67}$\\
MPSA-U & $\mathbf{-0.65 \pm 0.06}$ & $\mathbf{46.03 \pm 0.22}$ & $\mathbf{-07.59 \pm 0.06}$  & $\mathbf{-092.25 \pm 0.28}$ & $\mathbf{-092.73 \pm 2.00}$ & $\mathbf{+04.59 \pm 1.26}$\\
MPSA-D & $\mathbf{-0.65 \pm 0.06}$ & $\mathbf{46.03 \pm 0.22}$ & $\mathbf{-07.59 \pm 0.06}$  & $\mathbf{-093.11 \pm 0.41}$ & $\mathbf{-093.98 \pm 3.49}$ & $\mathbf{+04.90 \pm 2.30}$\\
GMM-F  & $-0.92 \pm 0.06$ & $05.37 \pm 0.23$  & $\mathbf{-07.60 \pm 0.06}$  & $-104.92 \pm 0.23$ & $-120.22 \pm 9.18$ & $-06.12 \pm 1.74$\\
GMM-S  & $-8.40 \pm 0.17$ & $15.11 \pm 1.01$ & $-11.67 \pm 0.14$ & $-100.02 \pm 0.28$ & $-099.67 \pm 1.18$ & $-17.15 \pm 0.59$\\
\botrule
\end{tabular*}
\end{footnotesize}
\end{table*}

In the first setting (MPSA $10$), all our MPSA models significantly outperform the classical GMMs. Since $p=10$ and $n=1000$, one probably has enough samples to estimate the full (anisotropic) covariance parameters, and the additional parsimony of our MPSA models yields a significantly higher penalized log-likelihood.
In the second setting (MPSA $100$), all our MPSA models again significantly outperform the classical GMMs. Since $p=100$ and $n=1000$, one probably does not have anymore enough samples to estimate the full covariance parameters, therefore the full GMM begins to severely drop in performance, while the spherical model is not flexible enough to model the data distribution due to its isotropic nature. In the third setting (Full $10$), since the true model is full and we have enough samples, the full GMM is the best estimator, and our MPSA models are on par with it, while the spherical GMM is below.
In the fourth setting (Full $100$), the true model is still full but we don't have enough samples anymore, so that the full GMM gets worse and the spherical GMM becomes better. Except for MPSA-R which is on par with GMM-S, our other MPSA models are significantly better. The same kind of observation goes for the fifth setting (Skew $100$).
In the sixth setting (Ionosphere), our MPSA models significantly outperform the other GMMs.

To conclude, our models achieve (almost always significantly) the best likelihood--parsimony tradeoffs in all data regimes, are robust to non-Gaussian distributions and real data, and even significantly outperform the full GMM when the true model is full and we have a small number of samples.
All our MPSA models are globally even in terms of final penalized log-likelihood, except for MPSA-R (the simplest and fastest model selection strategy) which seems to be slightly less adapted to high dimensional settings (Full $100$ and Skew $100$).

The previous experiments used the penalized log-likelihood~\eqref{eq:pen_lik} as a metric to evaluate the quality of the different GMMs. Beyond checking the monotonicity of the penalized log-likelihood over the CPEM iterations, these experiments showed that the MPSA models yield significantly higher penalized log-likelihoods than the classical GMMs---an empirical behavior that cannot be deduced from Theorem~\ref{thm:CPEM} alone. 
We believe that the penalized log-likelihood---as a tradeoff between goodness-of-fit and number of parameters---is a rather good indicator of the generalization properties of the learned models on unseen data, enjoying moreover statistical guarantees such as asymptotic consistency~\citep{keribin_consistent_2000,bai_consistency_2018}. That being said, it is clear that other interesting questions emanate from the study of the MPSA models, notably the one of density estimation: how well the MPSA models estimate the true underlying probability density. This is the purpose of the next subsection.

\subsection{Density estimation benchmark}\label{subsec:density_estimation}

\begin{figure*}
    \centering
    \includegraphics[width=.8\linewidth]{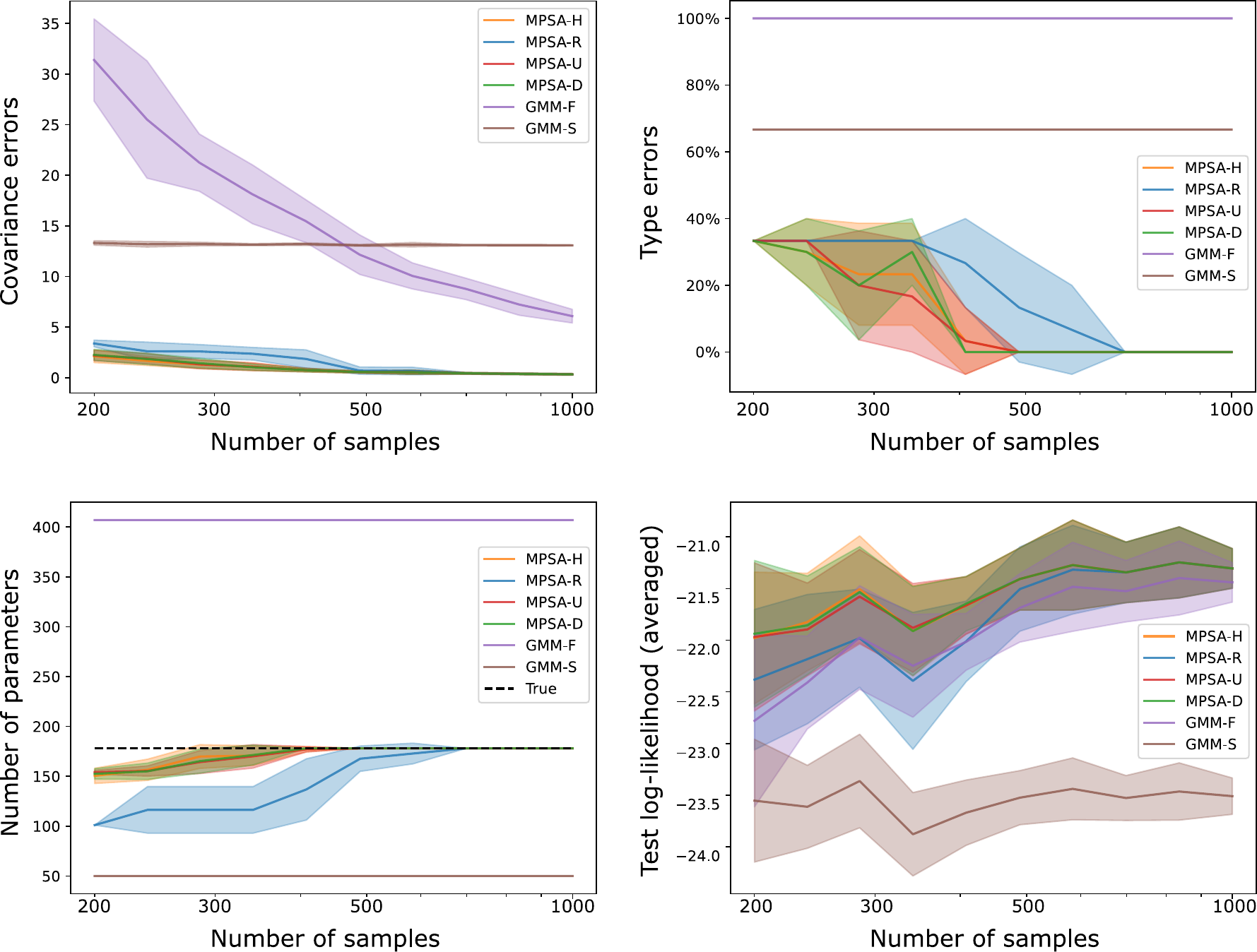}
    \caption{Density estimation with several GMMs. Several metrics are reported: the covariance errors, the type errors, the number of parameters and the (averaged) log-likelihood on an independent test set.}
    \label{fig:estimation}
\end{figure*}

We perform the following density estimation experiment. We generate data from the MPSA generative model, with varying $n$, $p=15$, $C=3$, ${(\gamma_1, \gamma_2, \gamma_3) = ((15), (5, 10), (5, 5, 5))}$, $(\pi_1, \pi_2, \pi_3) = (0.4, 0.3, 0.3)$, $\mu_c = 10$, $(\lambda_{11}, \lambda_{21}, \lambda_{31}) = (3, 2, 1)$, and $\lambda_{cp} / \lambda_{c1} = 0.1 \, \forall c$. The goal is to estimate the mixture parameters, that are the mixture proportions, means and covariances $(\pi_c,\mu_c,\Sigma_c)_{c=1}^C \in \Theta(\gamma)$, as well as the MPSA types $\gamma\in\mathcal{C}(p)^C$. For an increasing number of samples $n$, we display in Figure~\ref{fig:estimation} the covariance errors $\sum_{c=1}^C \|\hat\Sigma_c - \Sigma_c\|_F^2$, the type errors $1 - \sum_{c=1}^C \mathbf{1}_{\{\gamma_c\}}(\hat\gamma_c) / C$, the number of parameters $\kappa(\hat\gamma)$ and the (averaged) log-likelihood on an independent test set $\ln p(\, X^\mathrm{test} \mid \hat\theta \,) / n^\mathrm{test}$.\footnote{The Kullback-Leibler divergence---a classical density estimation metric---is not explicit for Gaussian mixture models~\citep{hershey_approximating_2007}.}
The component indexes $c$ of the true and estimated mixture densities are matched by minimizing the Euclidean distance between the centers; we check a posteriori that each estimated component is matched to one unique true component---i.e., the map corresponds to a permutation. Each metric is averaged over 10 independent experiments.

\begin{figure*}
    \centering
    \includegraphics[width=\linewidth]{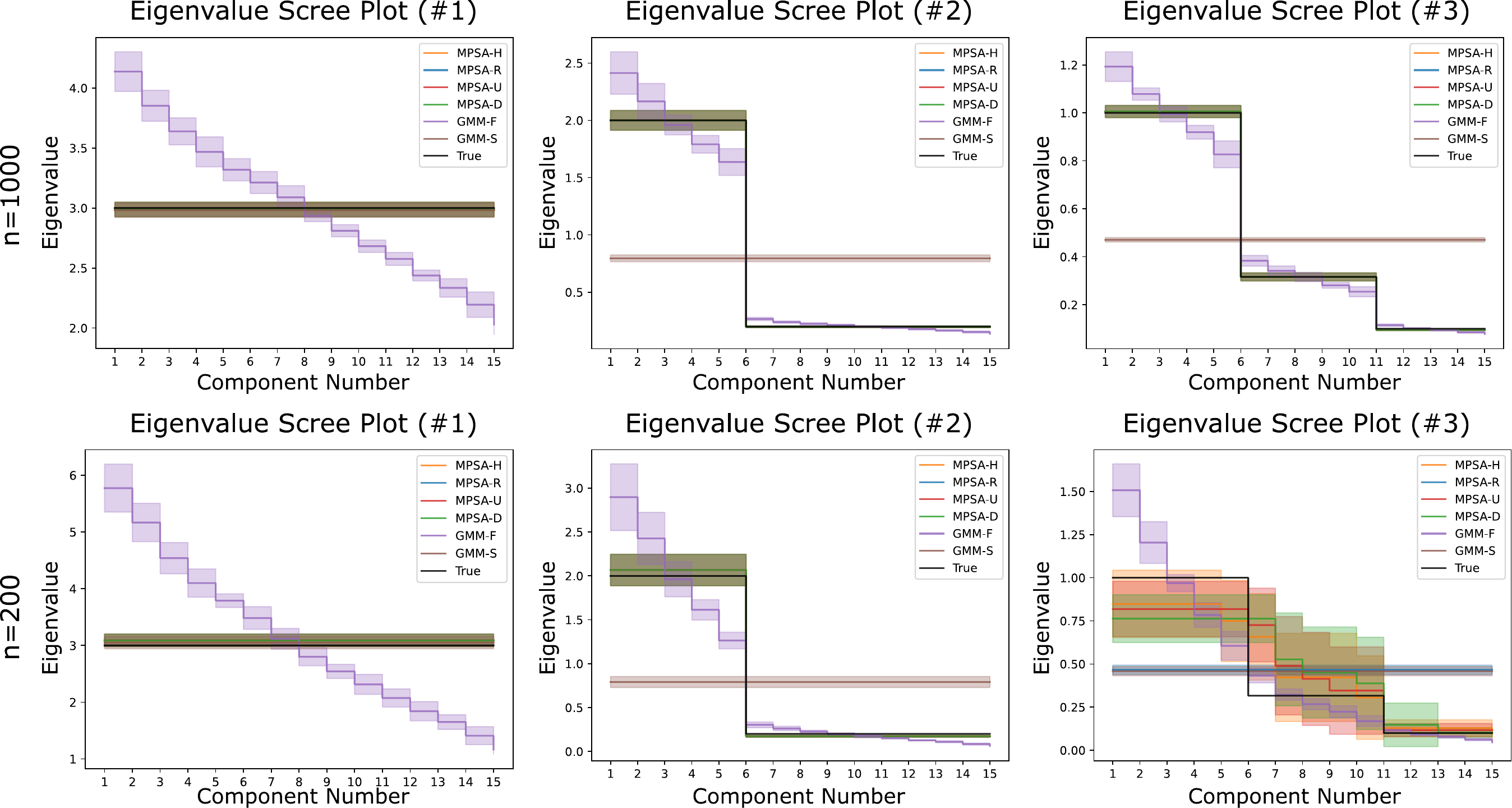}
    \caption{Covariance eigenvalue estimation with several GMMs for $n=1000$ (top) and $n=200$ (bottom). The three mixture components have, respectively, eigenvalue multiplicities $(15)$ (left), $(5, 10)$ (middle) and $(5, 5, 5)$ (right)---plotted in black. The four MPSA strategies cannot be visually distinguished on the first five plots, as they yield relatively similar results. On the sixth plot, MPSA-H/U/D can be distinguished, while MPSA-R cannot be distinguished from GMM-S.}
    \label{fig:eigenvalues}
\end{figure*}

To summarize the results, everything behaves as expected: the estimation errors decrease with the number of samples and tend to zero. Interestingly, the covariance estimation errors are significantly smaller for the MPSA models than for the spherical and full GMMs---even for $n=1000$. This can be explained by looking more closely at the covariance eigenvalue profiles, which is what we do in Figure~\ref{fig:eigenvalues}.
We see that the sample eigenvalues (represented by the purple curve) are highly dispersed around the population eigenvalues (represented by the black curve). This \textit{eigenvalue spreading} phenomenon is well known in high dimensions (see, for instance, \cite{johnstone_pca_2018}). In contrast, the eigenvalues estimated by the spherical GMM have low variance, but high bias for the second and third mixture component. The MPSA models not only achieve a significantly better estimation of the population eigenvalues, but also well capture the block structure, i.e., the eigenvalue multiplicities, which is not trivial. The eigenvalue plots of Figure~\ref{fig:eigenvalues} therefore well explain the results from Figure~\ref{fig:estimation} in terms of covariance and type estimation.

We can see moreover that the number of parameters increases with the number of samples and seems to converge to the accurate one, testifying of the BIC's asymptotic consistency. The spherical and full GMMs---respectively the most and least parsimonious GMMs belonging to the MPSA family---naturally bound the MPSA models on the plot of the number of parameters.

Finally, the test log-likelihood curves show that although the full GMM fails at estimating the covariance eigenvalues, it achieves similar density estimation results as the MPSA models, especially with respect to the spherical GMM.

Let us note that the estimation errors for the other mixture parameters---the proportions $\sum_{c=1}^C (\hat\pi_c - \pi_c)^2$ and the means $\sum_{c=1}^C \|\hat\mu_c - \mu_c\|_2^2$---were also evaluated: the errors also tend to 0 for all models. We did not display them since the differences between the GMMs were not significant. Indeed, since the clusters are well separated, the proportions and means are well estimated by all models, independently of their covariance structure.

The differences between the model selection strategies (H/R/U/D) are not stringent in terms of estimation, at least in the setting of the experiment and for $n$ large enough, where the parameters are well estimated for all strategies. The MPSA-R strategy seems overall slightly worse than the other strategies, as we can particularly see on the tasks of covariance, type and eigenvalue estimation.

\subsection{Clustering benchmark}
We now perform some clustering experiments, where the learning procedure is similar to the previous experiments, but the reported metric is the adjusted rand index (ARI)~\citep{rand_objective_1971,hubert_comparing_1985}, which is the classical substitute for the accuracy in clustering problems. This scores varies between (approximately) $0$ for a random clustering and (exactly) $1$ for a perfect clustering.
We add two competitors: HDDC~\citep{bouveyron_high-dimensional_2007} and spectral clustering~\citep{ng_spectral_2001}.
The first three datasets are synthetic. They all have in common that $C=3$, $(\pi_1, \pi_2, \pi_3) = (0.4, 0.3, 0.3)$, $(\lambda_{11}, \lambda_{21}, \lambda_{31}) = (10, 1, .1)$, $\lambda_{cp} / \lambda_{c1} = 0.01 \, \forall c$.
Now, more specifically, the first dataset (MPSA $10$) has $n=200$, $p=10$, $\mu_c \sim \mathcal{U}\lrp{\lrb{-1, 1}^p}$, and ${\gamma_1, \gamma_2, \gamma_3 = (1, p-1), (1, 2, p-3), (1, 2, 4, p-7)}$.
The second dataset (MPSA $50$) has the same characteristics as the previous one but with $\mu_c=0$ (mixed clusters, which is the case of the next ones), $p=50$ and $n=1000$.
The third one (Full $10$) has $p=10$, $n=200$ and ${(\gamma_1, \gamma_2, \gamma_3) = ((1^p), (1^p), (1^p))}$.\footnote{
Note that we choose particularly challenging settings (notably with close---or even exactly similar---centers) to show how our models can adapt the covariance structures, rather than the centers.
}
We then have three real datasets (Ionosphere~\citep{sigillito_ionosphere_1989}, Wine~\citep{aeberhard_wine_1991}, Breast~\citep{wolberg_breast_1995}), respectively with $(n, p, C) = (351, 34, 2), (178, 13, 3), (569, 30, 2)$.
The average clustering ARIs after 10 independent $n$-samples (or after a stratified $10$-fold cross-validation for the real datasets) are reported in Table~\ref{tab:clustering_ari}.
\begin{table*}[t]
\caption{
Comparison of the clustering ARIs ($\times 100$) reached by the MPSA-H/R/U/D models at the end of the CPEM algorithm, the standard GMMs, and the HDDC and spectral clustering models, for various challenging data distributions. ``$-$'' indicates an average ARI of $0$ or less, meaning that the clustering is similar or worse than random. The leading clustering ARIs are in bold.
}
\label{tab:clustering_ari}
\begin{tabular*}{\textwidth}{@{\extracolsep\fill}lcccccc}
\toprule%
Model & MPSA $10$ & MPSA $50$ & Full $10$ & Ionosphere & Wine & Breast\\
\midrule
MPSA-H   & $\mathbf{81 \pm 23}$ & $\mathbf{49 \pm 04}$ & $\mathbf{52 \pm 05}$ & $\mathbf{56 \pm 11}$ & $44 \pm 04$ & $\mathbf{80 \pm 06}$\\
MPSA-R   & $\mathbf{81 \pm 23}$ & $\mathbf{49 \pm 04}$ & $\mathbf{53 \pm 06}$ & $40 \pm 09$ & $42 \pm 05$ & $\mathbf{80 \pm 06}$\\
MPSA-U   & $\mathbf{81 \pm 23}$ & $\mathbf{48 \pm 04}$ & $\mathbf{50 \pm 05}$ & $36 \pm 06$ & $\mathbf{50 \pm 11}$ & $\mathbf{83 \pm 01}$\\
MPSA-D   & $\mathbf{81 \pm 24}$ & $\mathbf{50 \pm 04}$ & $\mathbf{49 \pm 05}$ & $\mathbf{50 \pm 12}$ & $\mathbf{51 \pm 12}$ & $\mathbf{79 \pm 06}$\\
GMM-F    & $\mathbf{80 \pm 24}$ & $\mathbf{48 \pm 02}$ & $\mathbf{48 \pm 03}$ & $32 \pm 06$ & $\mathbf{52 \pm 12}$ & $74 \pm 10$\\
GMM-S    & $37 \pm 17$ & $35 \pm 10$ & $43 \pm 11$ & $16 \pm 02$ & $39 \pm 02$ & $67 \pm 01$\\
HDDC     & $\mathbf{81 \pm 24}$ & $\mathbf{48 \pm 02}$ & $\mathbf{48 \pm 04}$ & $29 \pm 06$ & $04 \pm 04$ & $29 \pm 03$\\
Spectral & $60 \pm 31$ & $-$ & $-$ & $-$ & $-$ & $-$\\
\bottomrule
\end{tabular*}
\end{table*}

The results are obviously harder to comment than in the previous examples, since we evaluate the clustering ARI metric which is not the quantity that is optimized in the GMM learning algorithm~\eqref{eq:pen_lik}. Nonetheless, we have several interesting observations.
When the clusters centers are separated (MPSA $10$), all the GMMs achieve reasonable clustering except for the spherical one which does not fit well the anisotropic clusters, and the spectral clustering which is below in average. Note that the high standard deviation is probably due to a few instances of the 10 repetitions where the randomly-sampled centers are close, or where the randomly-sampled covariance orientations make the clusters harder to separate.
When the clusters are mixed (MPSA $50$ and Full $10$), the performances globally decrease since the setting is more complicated, and the spectral clustering even gets extremely bad (similar to random labeling). Our MPSA models are globally on par with the HDDC models on synthetic data (slightly better in average, but not significantly), although the HDDC models are run several times to tune the Cattell threshold hyperparameter. On real data, the MPSA models surprisingly yield better or even significantly better clustering performances.

The main conclusion of this experiment is that our MPSA models adapt well to several challenging synthetic clustering tasks, taking the most of the two extreme GMMs (full and spherical). 
A positive surprise is that our MPSA models compare more-than-well to the HDDC models (which they extend), although being fully unsupervised, while HDDC requires hyperparameter tuning. In a sense, our models are both robust and remove the need to determine the intrinsic dimensions.

\subsection{Running times}\label{subsec:runtimes}

As seen in the previous experiments, the four strategies are rather comparable in terms of likelihood--parsimony tradeoffs, density estimation and clustering, with the MPSA-R strategy being overall slightly worse than the other strategies. This subsection aims to compare the different strategies from the running time perspective.

\begin{figure}
    \centering
    \includegraphics[width=\linewidth]{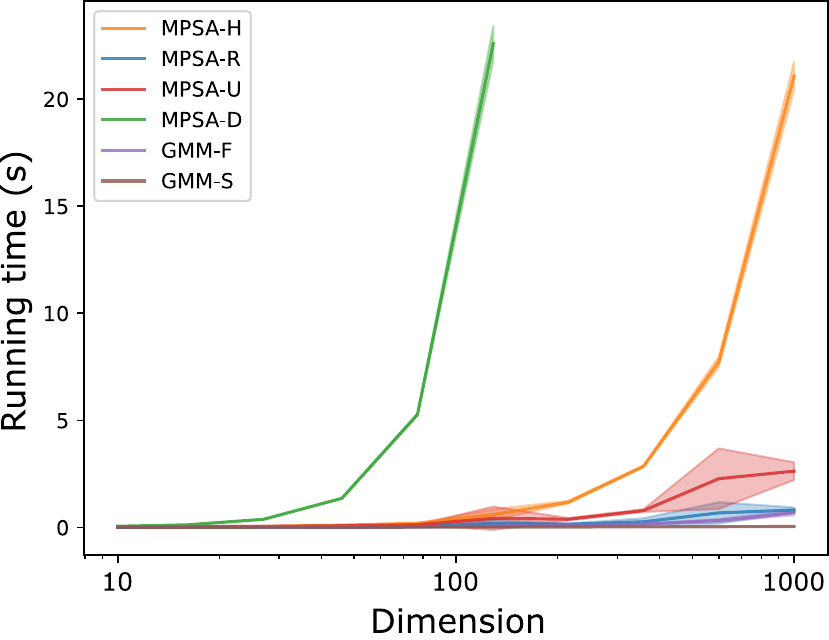}
    \caption{Running time benchmark for increasing dimension.}
    \label{fig:MPSA_times}
\end{figure}

We generate data from the MPSA generative model, with $n=1000$, $C=3$, $(\gamma_1, \gamma_2, \gamma_3) = ((1, p-1), (1, 2, p-3), (1, 2, 4, p-7))$, $(\pi_1, \pi_2, \pi_3) = (0.4, 0.3, 0.3)$, $\mu_c = 5$, $(\lambda_{11}, \lambda_{21}, \lambda_{31}) = (3, 2, 1)$, and $\lambda_{cp} / \lambda_{c1} = 0.01 \, \forall c$. The experiments are repeated independently 10 times. We display the running times of the different GMMs as a function of the dimension $p$ in Figure~\ref{fig:MPSA_times}.

The spherical GMM is significantly faster than all the other models---which makes sense since the estimation of the isotropic covariance matrices does not involve any matrix inversion or eigenvalue decomposition. The full GMM is faster than the MPSA models, in average, but it scales like the MPSA-R and MPSA-U strategies. The MPSA-H and MPSA-D strategies, for their part, are much slower, making them less adapted in high dimensions ($p \gtrapprox 1000$ for MPSA-H and $p \gtrapprox 100$ for MPSA-D). Let us now compare the four MPSA strategies.

MPSA-R is significantly faster than the other strategies---at the cost of apparently lower modeling performances, as we could see in the previous experiments. MPSA-U is the second fastest strategy, scaling like MPSA-R, although possibly requiring more iterations than MPSA-H, as seen in Figure~\ref{fig:clus_strategies}. MPSA-H is the third, as it involves a hierarchical clustering of the eigenvalues, which can become pretty costly in high dimensions (cf. Section~\ref{sec:PSA}). MPSA-D is the slowest---by far---as it starts from the full covariance model, and decreases the number of parameters gradually, which can take a lot of iterations to converge if the true model is parsimonious (cf. Figure~\ref{fig:clus_strategies}). Hence, the empirical running times rather match the previously built intuitions.

The comparisons between the MPSA and GMMs are to nuance, as our code is certainly not as much optimized as the one of scikit-learn~\citep{pedregosa_scikit-learn_2011}, from which the classical GMMs are drawn. Moreover, the current version of the code does not include certain numerical tricks such as the previously evoked truncated SVD, which is investigated in Section~\ref{subsec:TSVD}. We could expect well-engineered MPSA models to be significantly faster than full GMMs, as truncated SVDs are significantly faster than covariance inversion.

As a conclusion to the series of experiments performed in this section, we would probably recommend to adopt the MPSA-U strategy by default in general settings. Indeed, it is the second fastest strategy (scaling as the fastest one and being much faster than the third and fourth), and it has a good control on parsimony (as it is initialized with isotropic covariances and it increases gradually the number of parameters throughout the CPEM iterations, by finding the most informative gaps in the eigenvalue profiles). In contrast, the other strategies can possibly select overparameterized models in the first iterations, which can negatively impact learning in high dimensions.

\section{Application to single-image denoising}\label{sec:denoising}
Beyond traditional learning tasks like clustering, GMMs provide a flexible method for density modeling and therefore come with many other applications. In this section, we briefly describe the application of MPSA models to single-image denoising. More details about the denoising framework can be found in~\citet{houdard_high-dimensional_2018}.

\subsection{Image denoising with GMMs}\label{subsec:denoising_literature}
Many denoising methods rely on the \textit{self-similarity assumption}: any small patch extracted from an image has ``similar'' patches in the same image~\citep{efros_texture_1999,buades_non-local_2005}. This regularity and periodicity prior implies that patches extracted from natural images form well-defined clusters (see \cite{lebrun_secrets_2012} for more details). Such a redundancy enables to reduce the noise variance by averaging the similar patches. Under this self-similarity hypothesis, GMMs appear as natural models for the set of patches extracted from one image.

The celebrated work of~\cite{zoran_learning_2011} runs a GMM on a large database of natural image patches and use it for image denoising, deblurring and inpainting, outperforming all the traditional methods. However, this approach is not suitable for the task of \textit{single-image denoising}, where we just have one image as an input, that is moreover noisy; hence the need for parsimonious generative models. Consequently, \cite{houdard_high-dimensional_2018} propose to model the set of overlapping patches extracted from one image according to a low-dimensional MPPCA density with a shared mixture noise. The resulting model, HDMI (high-dimensional mixture models for unsupervised image denoising), automatically updates the intrinsic mixture dimensions $q_c$ during the EM based on a fixed hyperparameter $\sigma^2$ that models the variance of the image noise. More precisely, the intrinsic dimensions are chosen such that
\begin{equation}\label{eq:MS_noise}
 q_c = {\mathrm{argmin}}_{q \in [\![0, p[\![} \left| \sigma^2 -  \frac {1} {p - q} \sum_{j=q+1}^p \ell_{cj}\right|,
\end{equation}
where the $\ell_{cj}$ are the eigenvalues of the local responsibility-weighted covariance matrices $S_c$.
The noise level $\sigma^2$ can either be known (\textit{supervised denoising}) or unknown (\textit{unsupervised denoising}). In the latter case, it is estimated a posteriori as a hyperparameter, by minimizing the BIC of the learned mixture, similarly to HDDC~\citep{bouveyron_high-dimensional_2007}. The next paragraph describes the extension of HDMI with MPSA models.

Consider an image $X \in \R^{s_1\times s_2}$ that has been corrupted by some isotropic Gaussian noise $~{E \sim \N(0, \sigma^2_\mathrm{true} I_{s_1 s_2})}$. Then the observed image can be written as $\tilde X = X + E$.
We extract all the \textit{overlapping} $s\times s$ patches from $\tilde X$, from left to right, top to bottom. This yields a dataset $(\tilde x_i)_{i=1}^n$ of $n=(s_1-s+1)(s_2-s+1)$ noisy patches embedded in $\R^{p}$ with $p = s^2$. A nice illustration of such a dataset is provided in the Figure~1 of~\cite{houdard_high-dimensional_2018}. We assume that it follows an MPSA density~\eqref{eq:MPSA}, with the additional constraints $\lambda_{c d_c} = \sigma^2$,  $\forall c$. We run Algorithm~\ref{alg:CPEM} to infer the mixture parameters $\hat\theta$.
Proposition~\ref{prop:denoising} then gives an explicit and efficient formula to denoise the patches. 
\begin{proposition}[Denoising]\label{prop:denoising}
Let $\tilde{x}\in\R^{p}$ represent an $s\times s$ image patch following the noise model $\tilde{x} = x + \epsilon$, where $\epsilon \sim \N(0, \sigma^2 I_p)$ is an isotropic Gaussian noise and $x\in\R^p$ is the original patch---sampled from an MPSA model of types $\hat\gamma$ and parameters $(\dots \hat\pi_c, \hat\mu_c, \hat\Sigma_c - \sigma^2 I_p \dots)$.
Then the denoised patch $\hat x \coloneqq \mathbb{E}(\, x \mid \tilde x \,)$ can be expressed as
\begin{equation}\label{eq:denoising}
	\hat x = \sum_{c=1}^C w_c \biggl(\mu_c + \sum_{k=1}^{d_c - 1} \lrp{1 - \frac{\sigma^2}{\hat\lambda_{ck}}}\hat\Pi_{ck}(x - \mu_c)\biggr),    
\end{equation}
where $w_c = \mathbb{P}(\, y=c \mid \tilde x \,)$ (cf. Theorem~\ref{thm:E_step}).
\end{proposition}
\noindent Similarly to Section~\ref{subsec:complexity}, one can show that the use of the already-computed eigenvalues and eigenvectors and the substitution of the last eigenspace enables to switch from a computational complexity of $O(C n p^3)$ (with classical covariance inversion) to $O(C n p^2 q)$ (the most expensive step being the computation of $~{\hat\Pi_{ck} = \hat Q_{ck} {\hat {Q}_{ck}}\T}$).
The denoised patches are finally aggregated by performing a uniform reprojection---i.e., positioning back the denoised patches in the image from left to right, top to bottom and averaging them in the overlapping regions---yielding the denoised image $\hat X$.

\subsection{Denoising experiments}
We illustrate the previously presented approach on a $512 \times 512$ image with (unknown) noise-deviation $\sigma_\mathrm{true} = 30/255$ and patch size $~{s = 8}$.
We fit the MPSA-H/R/U models with $C \in \{3, 10, 20\}$ and compare them to the full, spherical, and (supervised) HDMI mixture models.\footnote{The HDMI models are learned in a supervised way (i.e. the true noise variance $\sigma^2_\mathrm{true}$ is given as an input), which is supposed to give a considerable advantage to HDMI over its competitors. We indeed found out that the results of HDMI were very sensitive with respect to the hyperparameter $\sigma^2$, and knowing that, we did not want to put arbitrary values in the grid search, which would have badly affected its performances.
}
We evaluate the quality of the denoising using the partial signal-to-noise ratio evaluation metric $~{\mathrm{PSNR} \coloneqq -10\log_{10}(\operatorname{MSE}(X, \hat X))}$, where MSE stands for \textit{mean squared error}, similarly to~\cite{houdard_high-dimensional_2018}.
We repeat the experiment 5 times independently and report the results in Table~\ref{tab:denoising}.
\begin{table*}[t]
\caption{Comparison of the denoising PSNR ($\uparrow$) reached by MPSA-H/R/U models at the end of the CPEM algorithm, with the ones of standard GMMs and the HDMI model, on the Simpson image, for different numbers of clusters $C$. The leading PSNRs are in bold.}
\label{tab:denoising}
\begin{tabular*}{\textwidth}{@{\extracolsep\fill}lccc}
\toprule
Model & $C=3$ & $C=10$ & $C=20$\\
\midrule
MPSA-H & $\mathbf{31.17 \pm 0.02}$ & $\mathbf{31.74 \pm 0.04}$ & $\mathbf{31.97 \pm 0.04}$\\
MPSA-R & $\mathbf{31.14 \pm 0.02}$ & $\mathbf{31.72 \pm 0.04}$ & $\mathbf{31.94 \pm 0.04}$\\
MPSA-U & $\mathbf{31.17 \pm 0.02}$ & $\mathbf{31.76 \pm 0.03}$ & $\mathbf{31.98 \pm 0.03}$\\
HDMI($\sigma^2_\mathrm{true}$) & $\mathbf{31.11 \pm 0.01}$ & $\mathbf{31.72 \pm 0.03}$ & $\mathbf{31.90 \pm 0.02}$\\
GMM-F  & $30.73 \pm 0.03$ & $30.27 \pm 0.05$ & $29.33 \pm 0.03$\\
GMM-S  & $22.71 \pm 0.01$ & $25.76 \pm 0.13$ & $27.56 \pm 0.04$\\
\bottomrule
\end{tabular*}
\end{table*}

We see that the MPSA-H/R/U models always achieve significantly greater PSNR than the classical GMMs. Moreover, while the denoising performance of the MPSA and GMM-S models increase with the number of patches, the ones of GMM-F surprisingly decrease. This can be explained by the fact that increasing the number of clusters reduces the number of patches-per-cluster, so that the full GMM becomes overparameterized, resulting in poorer modeling.
The supervised HDMI model is on par or even significantly worse than the MPSA models, which is a good surprise since the MPSA models do not have access to the noise level.

\begin{figure*}
\begin{center}
\centerline{\includegraphics[width=\linewidth]{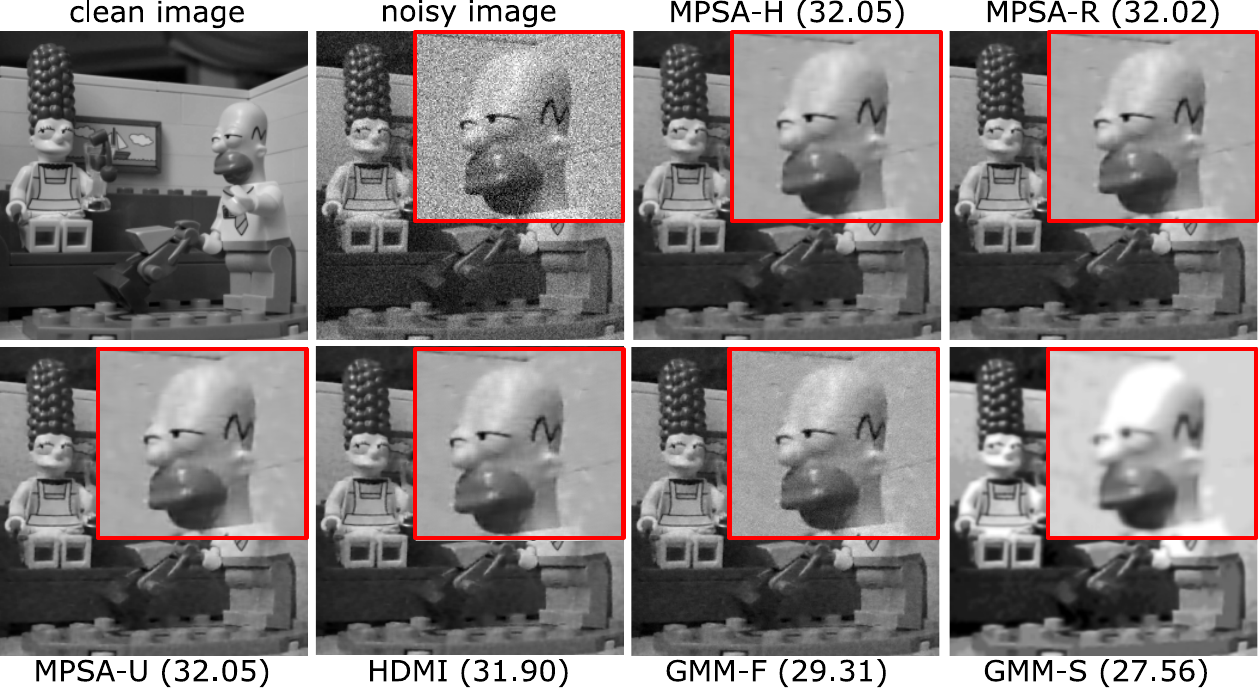}}
\caption{Denoised images and PSNR with different parsimonious GMMs with $C=20$.}
\label{fig:denoising}
\end{center}
\end{figure*}

Moreover, we display in Figure~\ref{fig:denoising} the denoised images for the fifth independent repetition of the $C=20$ experiment.
We observe that the GMM-S denoised image is relatively blurry with respect to the other methods, and not much textured. This can be explained by the simplicity of the underlying model for the clusters, which is isotropic Gaussian, so that the denoised patches are essentially sparse convex combinations of a few prototypes.\footnote{The denoising formula~\eqref{eq:denoising} applied in the context of a spherical GMM indeed yields $\hat x = \sum_{c=1}^C w_c \mu_c$.} 
In contrast, the full GMM has too much degrees of freedom, so that it ``overfits'' the noise.\footnote{Examining the denoising formula~\eqref{eq:denoising}, we see that if the estimated noise $\sigma^2$---consisting in the proportion-weighted average of the smallest eigenvalues of the $S_c$---is small with respect to the other eigenvalues of the $S_c$, then one roughly has $\hat x \approx \tilde x$.
}
The MPSA models stand in the middle and achieve better denoising, both qualitatively (visually) and quantitatively (via the PSNR metric). 

\begin{figure*}
\begin{center}
\centerline{\includegraphics[width=\linewidth]{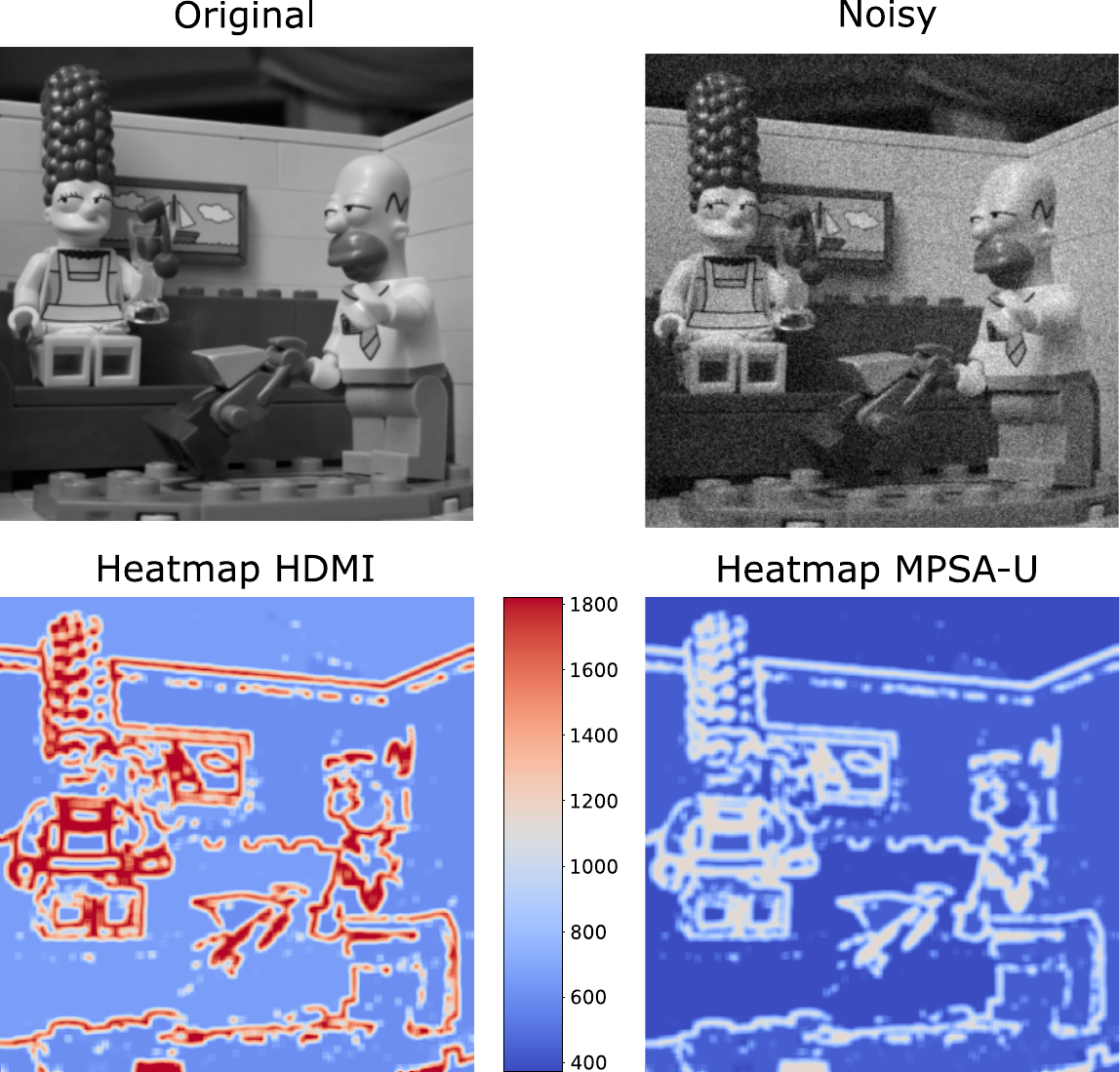}}
\caption{Heatmap-based representation of the clustering achieved by the HDMI and MPSA-U models on the patches of the noisy Simpson image, for $C=3$. The patches are colored according to the number of parameters of the mixture component they belong to.}
\label{fig:denoising_colors}
\end{center}
\end{figure*}

Finally, we propose in Figure~\ref{fig:denoising_colors} a heatmap-like visualization of the clusters directly on the image, for both (supervised) HDMI and MPSA-U models with $C=3$. More precisely, we map each patch to the cluster it most likely belongs to ($\hat c = \argmax_{c \in [\![1, C]\!]} \mathbb{P}(\, y=c \mid \tilde x \,)$), and color this patch according to the number of parameters of the underlying mixture component (which is a PPCA model for HDMI and a PSA model for MPSA-U).
We see that the clusters well recover the different textures that are present in the image, the more parsimonious being related to uniform textures (like walls) and the more complex being related to more contrasted textures (like edges and boundaries). 
Moreover, we see that the HDMI model requires much more parameters to model the patches (mostly for the textured zones but also for the uniform zones). More precisely, one has:
\begin{align*}
    \gamma_{\mathrm{HDMI}} &\coloneqq  ((1^9, 55), (1^{10}, 54), (1^{39}, 25)),\\
    \gamma_{\mathrm{MPSA}} &\coloneqq  ((1^5, 59), (1^6, 58), (1^9, 2, 1, 4, 2, 2, 44)),
\end{align*}
which yields by virtue of Proposition~\ref{prop:MPSA_kappa}: 
\begin{align*}
    \kappa(\gamma_{\mathrm{HDMI}}) &= 2 + 605 + 660 + 1820 = 3087,\\
    \kappa(\gamma_{\mathrm{MPSA}}) &= 2 + 375 + 434 + 1140 = 1951.
\end{align*}
This means that the uniform textures are attributed by HDMI an intrinsic dimension of respectively $9$ and $10$, while MPSA-U only requires respectively $5$ and $6$ dimensions. In contrast, the complex textures have an intrinsic dimension of $39$ with the HDMI model, while they have an intrinsic dimension of $20(=64-44)$ with the MPSA-U model.\footnote{
A legitimate question might be: does an HDMI model of similar complexity as the MPSA-U model yield smaller PSNRs? The only way to reduce the number of parameters of the HDMI model being to increase the estimated noise (cf. Eq.~\eqref{eq:MS_noise}), we tried several values of $\sigma$. For $\sigma = 40/255$, one gets the types $(1, 63), (1, 63), (1, 63)$ and the PSNR $26.8$; for $\sigma = 35/255$, one gets the types $(1, 63), (1, 63), (1^2, 62)$ and the PSNR $27.8$; for $\sigma = 32/255$, one gets the types $(1, 63), (1, 63), (1^7, 57)$ and the PSNR $30.7$; for $\sigma = 30/255 (=\sigma_\mathrm{true})$, one gets the types $(1^9, 55), (1^{10}, 54), (1^{39}, 25)$ and the PSNR $31.1$.
We notice an extreme sensitivity of the selected model complexity and PSNRs with respect to the choice of hyperparameter $\sigma$, but it seems that with similar numbers of parameters, HDMI and MPSA-U behave relatively similarly.
}
Finally, we find that the heatmap obtained with MPSA-U is cleaner visually than the one of HDMI, probably due to the added parsimony.

\begin{remark}[Influence of $s$ and $C$ on denoising]
    In all the experiments of this section, the patch size is set to $s=8$  and the number of components $C$ varies heuristically according to the denoising task. We discuss further in depth the influence of $s$ and $C$ on the denoising performances and running times in Section~\ref{subsec:denoising_influence}. Similar experiments can be found in~\cite{houdard_high-dimensional_2018}.
\end{remark}

\section{Conclusion}
We introduced a new family of parsimonious Gaussian mixture models with piecewise-constant covariance eigenvalue profiles. These were motivated by the important parsimony achieved by equalizing similar eigenvalues~\citep{szwagier_curse_2025}, going beyond classical low-rank Gaussian mixture models~\citep{tipping_mixtures_1999,bouveyron_high-dimensional_2007}. While the unsupervised learning of the mixture parameters was easily enabled via an EM algorithm, the automatic inference of the groups of eigenvalues to equalize was more technical. It was finally achieved via a \textit{provably-monotonous componentwise penalized EM} algorithm. The numerical experiments showed the superior likelihood--parsimony tradeoffs achieved by our models when compared to the two extreme cases: the full GMM (with distinct covariance eigenvalues) and the spherical GMM (with equal covariance eigenvalues), and the effective application of our models to clustering, density modeling and denoising tasks.

Beyond our extensive experiments to analyze the learning algorithm, one could investigate more data regimes, variants and applications. We notably believe that our new CPEM algorithm has the potential to be applied to many classical low-rank mixture models, removing the need for post-training hyperparameter selection.
Instead of an EM algorithm, one could also directly optimize the penalized log-likelihood with gradient descent techniques~\citep{hosseini_alternative_2020}, drawing from the recent relaxation of parsimony of~\cite{szwagier_eigengap_2025}.
The MPSA models could also be naturally applied to the numerous generative modeling tasks for which classical GMMs are often used. 
Their expressivity could be improved by drawing from the recent advances in GMMs and machine learning~\citep{richardson_gans_2018,loconte_subtractive_2023,byrski_gaussian_2024}. Robustness could be achieved by integrating our principles into the recent mixtures of high-dimensional elliptical distributions of~\citet{oudoumanessah_scalable_2024}.

Other potential applications of our parsimonious models are \textit{manifold learning} and \textit{surface reconstruction}. Indeed, it is often assumed that the observed data has been sampled from a low-dimensional manifold (up to isotropic Gaussian noise). The density could be fit with our MPSA models, and the underlying shape could be recovered with classical surface reconstruction methods. The types of the mixture components could tell something about the local intrinsic dimensions of the underlying shape~\citep{akhoj_principal_2023,buet_flagfolds_2023}.

\begin{figure*}
\begin{center}
\centerline{\includegraphics[width=\linewidth]{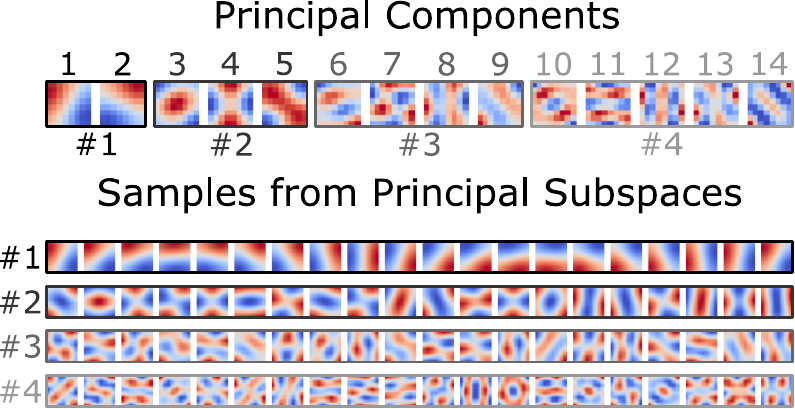}}
\caption{Illustration of the principal subspaces found by fitting an MPSA-U model to patches extracted from the clean Simpson image. The inferred eigenvalue multiplicities for the first component are $\gamma_1 = (2, 3, 4, 5, \dots)$, which means that the principal subspaces $\#1, \#2, \#3, \#4$ have respective dimensions $2$, $3$, $4$ and $5$. We display some samples from these principal subspaces (more precisely, convex combinations of the associated principal components). These can be interpreted as feature subspaces with (limited) frequency invariance, which are reminiscent of the findings of~\cite{szwagier_curse_2025} about the curse of isotropy.
}
\label{fig:patches}
\end{center}
\end{figure*}

Finally, since the PSA models have been originally introduced to improve the interpretation of principal components with close eigenvalues, one should definitely investigate the interpretation of the principal subspaces found by the MPSA models. We conclude this paper with a simple example on image patches. 
As discussed and experimented in Section~\ref{sec:denoising}, image patches are classically represented with Gaussian mixtures, and we believe that they carry some invariances that could be well modeled with piecewise-constant eigenvalue profiles. Indeed, many classical works have noticed that the principal components in natural image patch datasets have repeated eigenvalues, accounting for shared frequencies (see, for instance, \citet[FIG.~1]{olshausen_emergence_1996}, \citet[Figure~5(a)]{mairal_online_2010} and \citet[Figure~6]{zoran_learning_2011}). Therefore, the MPSA models could reveal these hidden symmetries by detecting the multiple covariance eigenvalues in the mixture components, similarly to~\cite{szwagier_curse_2025}. 
We illustrate this perspective by fitting an MPSA-U model with $C=3$ on patches extracted from the \textit{clean} Simpson image. The output model types are quite interesting, e.g. $\gamma = (2, 3, 4, 5, \dots)$ for the first cluster. We display in Figure~\ref{fig:patches} the first few principal components associated with the first cluster, gathered according to their eigenvalue multiplicities, and then sample some points from the associated principal subspaces. This essentially boils down to generating convex combinations of the principal components with similar eigenvalues.
We observe some interesting patterns within the principal subspaces, which can somewhat be interpreted as feature subspaces with (limited) frequency invariance.

\backmatter





\bmhead{Acknowledgments}

This work was supported by the ERC grant \#786854 G-Statistics from the European Research Council under the European Union’s Horizon 2020 research and innovation program and by the French government through the 3IA Côte d’Azur Investments ANR-23-IACL-0001 managed by the National Research Agency.





\begin{appendices}

\section{Proofs}\label{app:proofs}
This section provides the proofs of the propositions and theorems from the main text.
\subsection{Proposition~\ref{prop:MPSA_kappa}: number of parameters for the MPSA models}
\begin{proof}
The parameters of the MPSA model of types $\gamma$ are $\pi_1, \mu_1, \Sigma_1, \dots, \pi_C, \mu_C, \Sigma_C$, with the constraints $\pi_c \in \R_{\geq 0}, \sum_{c=1}^C \pi_c = 1$, $\mu_c \in \R^p$ and $\Sigma_c \in \Sym_+(\gamma_c)$.
The number of free parameters for $\pi_1, \dots, \pi_C$ (belonging to the \textit{standard simplex}) is $C-1$, since the last mixture proportion $\pi_C$ is automatically determined from the $C-1$ other ones, via $\pi_C = 1 - \sum_{c=1}^{C-1} \pi_c$.
Then for each $c \in [\![1, C]\!]$, the parameters $\mu_c$ and $\Sigma_c$ describe a PSA model of type $\gamma_c$. Therefore, the number of free parameters is $\kappa(\gamma_c)$, as defined in Equation~\eqref{eq:PSA_kappa}. Since neither the means nor the covariance matrices are shared between the mixture components, the number of parameters for $\mu_1, \Sigma_1, \dots, \mu_C, \Sigma_C$ is simply the sum $\sum_{c=1}^C \kappa(\gamma_c)$.
Eventually, since the mixture proportions are independent from the means and covariance matrices, the total number of parameters is ${\kappa(\gamma) = C - 1 + \sum_{c=1}^C \kappa(\gamma_c)}$.
\end{proof}

\subsection{Theorem~\ref{thm:E_step}: E-step of the EM}
\begin{proof}
The posterior log-probabilities can be rewritten using Bayes' theorem:
\begin{multline}\label{eq:posterior}
    \ln\mathbb{P}(\,y=c \mid x, \theta\,) = \ln\lrp{\pi_c \, \mathcal{N}(\, x \mid \mu_c, \Sigma_c\,)}\\
    - \ln\lrp{\sum_{c'=1}^C \pi_{c'} \, \mathcal{N}(\,  x \mid \mu_{c'}, \Sigma_{c'}\,)}.
\end{multline}
Let us now develop the operand using the eigendecomposition of the sample covariance matrices, $~{\Sigma_c = \sum_{k=1}^{d_c} \lambda_{ck} \Pi_{ck}}$, where $\lambda_{ck}$ are the (ordered decreasing) eigenvalues of $\Sigma_c$ and $\Pi_{ck}$ are the associated eigenspaces, of dimension $\gamma_{ck}$.
One has $~{\ln| \Sigma_c| = \sum_{k=1}^{d_c} \gamma_{ck}\ln{\lambda_{ck}}}$, and via classical trace tricks, one gets 
\begin{align}
    \operatorname{d}^2_{\Sigma_c}(x, \mu_c) &\coloneqq (x-\mu_c)\T\Sigma_c^{-1}(x-\mu_c),\\ 
    &= \tr{\Sigma_c^{-1}(x-\mu_c)(x-\mu_c)\T},\\
    &= \sum_{k=1}^{d_c} \tr{\lambda_{ck}^{-1} \Pi_{ck}(x-\mu_c)(x-\mu_c)\T},\\
    &= \sum_{k=1}^{d_c}\frac{(x - \mu_c)\T \Pi_{ck} (x - \mu_c)}{\lambda_{ck}}.
\end{align}
Consequently, one has 
\begin{multline}
    \ln(\pi_c \, \mathcal{N}(\,  x \mid \mu_c, \Sigma_c \,)) = \ln\pi_c - \frac{p}{2}\ln 2\pi\\
    - \frac{1}{2} \sum_{k=1}^{d_c} \lrp{\gamma_{ck}\ln{\lambda_{ck}}+ \frac{(x - \mu_c)\T \Pi_{ck} (x - \mu_c)}{\lambda_{ck}}}.
\end{multline}
Let us define the cost function $K_c$ similarly to~\cite{bouveyron_high-dimensional_2007} via the following relationship: $\ln(\pi_c \, \mathcal{N}(\,  x \mid \mu_c, \Sigma_c \,)) = -\frac{K_c(x) + p \ln 2\pi}{2}$. 
Re-injecting $K_c$ in the posterior probabilities yields
\begin{align}
	\mathbb{P}(\,y=c \mid x, \theta\,) &= \frac{\exp\lrp{- \frac{K_c(x) + p \ln 2\pi}{2}}}{\sum_{c'=1}^C \exp\lrp{-\frac{K_{c'}(x) + p \ln 2\pi}{2}}},\\
	&= \frac{1}{\sum_{c'=1}^C \exp\lrp{\frac{K_c(x) - K_{c'}(x)}2}}.
\end{align}
Re-injecting $K_c$ in the log-likelihood yields
\begin{align}
\operatorname{LL}(\theta) &= \sum_{i=1}^n \ln\lrp{\sum_{c=1}^C \pi_c \, \mathcal{N}( \, x_i \mid \mu_c, \Sigma_c \,)},\\
	&= \sum_{i=1}^n \ln\lrp{\sum_{c=1}^C \exp\lrp{-\frac{K_c(x_i) + p \ln 2\pi}{2}}}.
\end{align}
\begin{remark}
    The second term in~\eqref{eq:posterior} is independent of $c \in [\![1, C]\!]$. Therefore, only the first term intervenes in the cluster assignment problem, which can be rewritten as $~{y = \argmax_{c \in [\![1, C]\!]} \ln\lrp{\pi_c \, \mathcal{N}(\,  x \mid \mu_c, \Sigma_c \,)}}$. 
    Then one has $y = \argmin_{c \in [\![1, C]\!]}\, K_c(x)$, which makes the implementation more efficient.
\end{remark}
\end{proof}

\subsection{Theorem~\ref{thm:M_step}: M-step of the EM}
\begin{proof}
The optimal mixture proportions are
\begin{equation}
    \hat\pi_1, \dots, \hat\pi_C = \argmax_{\substack{\pi_c \geq 0\\ \sum_{c=1}^C \pi_c = 1}} \sum_{c=1}^C \lrp{\ln\pi_c \sum_{i=1}^n  t_{ci}},
\end{equation}
whose solutions are given by $\hat\pi_c = \frac1n \sum_{i=1}^n t_{ci}$.
For each $c \in [\![1, C]\!]$, let us write $~{S_c \coloneqq \frac1{n \hat \pi_c} \sum_{i=1}^n t_{ci} \lrp{x_i - \mu_c} \lrp{x_i - \mu_c}\T \in \Sym(p)}$. Then one can rewrite the objective function as
\begin{multline}
    \mathbb{E}_{\,y\mid x, \theta\,} [\ln\mathcal{L}_{\mathrm{c}}((\hat\pi_c, \mu_c, \Sigma_c)_{c=1}^C)] = \text{const}\\
    - \frac{1}{2} \sum_{c=1}^C n\hat\pi_c \lrp{\ln|\Sigma_c| + \tr{ \Sigma_c^{-1} S_c}}.
\end{multline}
Consequently, the M-step on the mean and covariance parameters can be decomposed into the following independent minimization problems for $c \in [\![1, C]\!]$:
\begin{equation}
    \hat\mu_c, \hat\Sigma_c = \argmin_{\substack{\mu_c\in\R^p\\ \Sigma_c\in\Sym_+(\gamma_c)}} \,  \ln|\Sigma_c|  + \tr{ \Sigma_c^{-1} S_c}.
\end{equation}
The solutions to such problems are given in the maximum likelihood estimation theorem of~\cite{szwagier_curse_2025}. One has
$\hat\mu_c = \frac1{n \hat \pi_c} \sum_{i=1}^n t_{ci} \, x_i$ and $\hat\Sigma_c = \sum_{k=1}^{d_c} \hat\lambda_{ck}(t_c, \gamma_c) \, \hat \Pi_{ck}(t_c, \gamma_c)$, where 
$\hat\lambda_{ck}$ and $\hat \Pi_{ck}$
are the block-averaged eigenvalues and eigenspaces of $S_c$.
\end{proof}

\subsection{Theorem~\ref{thm:CPEM}: monotonicity of the CPEM}
\begin{proof}
Let $\mathbb{E}_{\,y\mid x,\theta\,} [\ln\mathcal{L}_{\mathrm{c}}(\cdot)]$ be the {expected} complete-data log-likelihood~\eqref{eq:expected_ll}. Then one has
\begin{equation}
    \mathbb{E}_{\,y\mid x,\theta\,} [\ln\mathcal{L}_{\mathrm{c}}(\theta(\gamma'))] = \sum_{c=1}^C \Psi_c(\gamma_c') + \text{const},
\end{equation}
and 
\begin{equation}
    \kappa(\gamma') = \sum_{c=1}^C \kappa(\gamma_c') + \text{const}.
\end{equation}
Therefore, summing the inequality~\eqref{eq:component_increase} from the theorem statement over $c \in [\![1, C]\!]$ and adding the constants yields
\begin{multline}\label{eq:CPEM_ineq_1}
    \mathbb{E}_{\,y\mid x,\theta\,} [\ln\mathcal{L}_{\mathrm{c}}(\theta(\gamma'))] - \alpha \, \kappa(\gamma')\\ 
    \geq
    \mathbb{E}_{\, y \mid  x, \theta\, } [\ln\mathcal{L}_{\mathrm{c}}(\theta(\gamma))] - \alpha \, \kappa(\gamma),
\end{multline}
where $~{\theta(\gamma) \coloneqq (\hat\pi_c(t_c), \hat\mu_c(t_c), \hat\Sigma_c(t_c, \gamma_c))_{c}\in\Theta(\gamma)}$.
Additionally, by definition of $\theta(\gamma)$, that is $\theta(\gamma) \coloneqq \argmax_{\theta'\in\Theta(\gamma)}\mathbb{E}_{\, y \mid  x, \theta\, } [\ln\mathcal{L}_{\mathrm{c}}(\theta')]$, 
one has
\begin{multline}\label{eq:CPEM_ineq_2}
    \mathbb{E}_{\, y \mid  x, \theta\, } [\ln\mathcal{L}_{\mathrm{c}}(\theta(\gamma))] - \alpha \, \kappa(\gamma)\\ \geq
    \mathbb{E}_{\, y \mid  x, \theta\, } [\ln\mathcal{L}_{\mathrm{c}}(\theta)] - \alpha \, \kappa(\gamma).
\end{multline}
The two previous equations (\ref{eq:CPEM_ineq_1}, \ref{eq:CPEM_ineq_2}) simply mean that a classical EM iteration for the MPSA model of types $\gamma$ increases the expected complete-data penalized log-likelihood $\mathbb{E}_{\, y \mid  x, \theta\, }  [\ln\mathcal{L}_{\mathrm{c}}(\cdot)] - \alpha \, \kappa(\gamma)$, and that an adapted change of types from $\gamma$ to $\gamma'$ increases this quantity even more.
Using Gibbs' inequality as classically done in the proof of the EM monotonicity~\citep{dempster_maximum_1977}, one gets
\begin{multline}
    \ln \mathcal{L}(\theta(\gamma')) - \ln \mathcal{L}(\theta)\\
    \geq \mathbb{E}_{\, y \mid  x, \theta\, } [\ln\mathcal{L}_{\mathrm{c}}(\theta(\gamma'))] - \mathbb{E}_{\, y \mid  x, \theta\, } [\ln\mathcal{L}_{\mathrm{c}}(\theta)].
\end{multline}
We conclude the proof by subtracting $\alpha(\kappa(\gamma') - \kappa(\gamma))$ on both sides and using (\ref{eq:CPEM_ineq_1}, \ref{eq:CPEM_ineq_2}):
\begin{multline}\label{eq:gibbs}
    (\ln \mathcal{L}(\theta(\gamma')) - \alpha \, \kappa(\gamma')) - (\ln \mathcal{L}(\theta) - \alpha \, \kappa(\gamma))\geq\\
    (\mathbb{E}_{\, y \mid  x, \theta\, } [\ln\mathcal{L}_{\mathrm{c}}(\theta(\gamma'))] - \alpha \, \kappa(\gamma'))\\
    - (\mathbb{E}_{\, y \mid  x, \theta\, } [\ln\mathcal{L}_{\mathrm{c}}(\theta)]  - \alpha \, \kappa(\gamma)) \geq 0.
\end{multline}
\end{proof}

\subsection{Proposition~\ref{prop:denoising}: GMM-based single-image denoising}
\begin{proof}
The Proposition~1 of~\cite{houdard_high-dimensional_2018} proves that $\mathbb{E}(\,  x \mid \tilde x\,) = \sum_{c=1}^C w_c \psi_c(\tilde x) $ with $~{\psi_c(\tilde x) \coloneqq \mu_c + (I_p - \sigma^2 {\Sigma_c}^{-1})(\tilde x - \mu_c)}$. Then, using the eigendecomposition of the sample covariance matrices, $\Sigma_c = \sum_{k=1}^{d_c} \lambda_{ck} \Pi_{ck}$, where $\lambda_{ck}$ are the (ordered decreasing) eigenvalues of $\Sigma_c$ and $\Pi_{ck}$ are the associated eigenspaces, one can rewrite:
\begin{align}
	\psi_c(\tilde x) &= {\mu_c + \lrp{I_p - \sigma^2 \sum_{k=1}^{d_c} \lambda_{ck}^{-1}\Pi_{ck}}(\tilde x - \mu_c)}\\
	&= {\mu_c + \lrp{\sum_{k=1}^{d_c} \Pi_{ck} - \sigma^2 \sum_{k=1}^{d_c} \lambda_{ck}^{-1}\Pi_{ck}}(\tilde x - \mu_c)},\\
	&= {\mu_c + \sum_{k=1}^{d_c} \lrp{1 - \frac{\sigma^2}{\lambda_{ck}}}\Pi_{ck}(\tilde x - \mu_c)},\\
	&= {\mu_c + \sum_{k=1}^{d_c - 1} \lrp{1 - \frac{\sigma^2}{\lambda_{ck}}}\Pi_{ck}(\tilde x - \mu_c)}.
\end{align}
Let us insist on the last line, which is due to the constraints $\lambda_{cd_c} = \sigma^2$ for all $c \in [\![1, C]\!]$ and which shows that the patches are actually projected to the low-dimensional subspaces $~{\mathcal{V}_c \coloneqq \operatorname{Span}(\sum_{k=1}^{d_c - 1} \Pi_{ck})}$. This represents an important computational gain in high dimensions (from $O(p^3)$ to $O(p^2)$) for the same reasons as outlined in Section~\ref{subsec:complexity}.
\end{proof}

\section{Experiments}\label{app:exp}

\subsection{Penalty factor $\alpha$}\label{subsec:alpha}

In all the experiments, we set the penalty factor to $\alpha = \ln(n)/2$. This corresponds to a BIC-like penalized log-likelihood~\eqref{eq:pen_lik}, which therefore enjoys theoretical foundations and asymptotic consistency guarantees~\citep{schwarz_estimating_1978,keribin_consistent_2000,bai_consistency_2018}. Notably, as we can see it in the density estimation experiments of Section~\ref{subsec:density_estimation}, the estimated mixture parameters converge to the true ones when the number of samples goes to infinity. We also chose this criterion because this is a very classical one in model-based clustering (see, for instance, \citet{scrucca_mclust_2016,bouveyron_model-based_2019}).

Alternatives criteria could be the Akaike information criterion (AIC)~\citep{akaike_new_1974} or the integrated completed likelihood (ICL)~\citep{biernacki_assessing_2000}. These criteria might not enjoy the same theoretical guarantees and result in different empirical results. Notably, the AIC ($\alpha = 1$) is known to favor overparameterized models with respect to the BIC ($\alpha = \ln(n)/2$). While AIC and BIC directly express under the form of Equation~\eqref{eq:pen_lik}---i.e., as a log-likelihood penalized by the number of samples---ICL has a slightly different expression, which makes it non-integratable in the current framework, especially if we want monotonicity guarantees such as the ones of Theorem~\ref{thm:CPEM}.

\begin{figure}
    \centering
    \includegraphics[width=\linewidth]{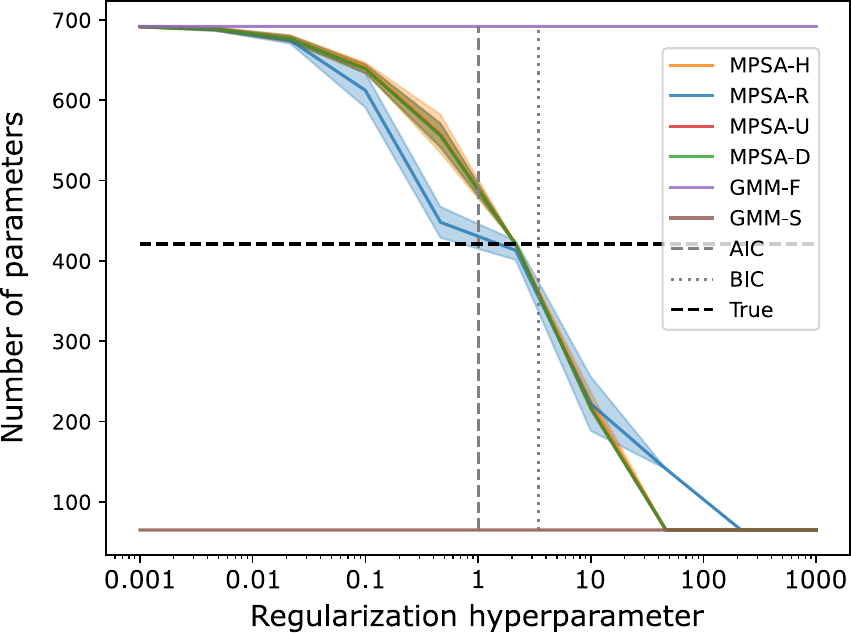}
    \caption{Influence of the regularization hyperparameter $\alpha$ on the number of parameters of the inferred mixture models. The true number of parameters is $\kappa(\gamma)=421$.}
    \label{fig:alpha}
\end{figure}

To account for the influence of the penalty factor, we report in Figure~\ref{fig:alpha} the evolution of the number of parameters of the learned mixture model (via the CPEM algorithm) as a function of $\alpha$.\footnote{The experimental details are the following ones: the data is generated from an MPSA density with $n=1000$, $p=20$, $C=3$, ${(\gamma_1, \gamma_2, \gamma_3) = ((5, 15), (5, 5, 10), (5, 5, 5, 5))}$, $(\pi_1, \pi_2, \pi_3) = (0.4, 0.3, 0.3)$, $\mu_c = 10$, $(\lambda_{11}, \lambda_{21}, \lambda_{31}) = (3, 2, 1)$, and $\lambda_{cp} / \lambda_{c1} = 0.1 \, \forall c$. The number of parameters is therefore $\kappa(\gamma)=421$. The experiments are repeated independently 10 times.}
We see that the number of parameters indeed decreases with the penalty factor $\alpha$, going from the highest complexity (corresponding to the full GMM) to the lowest one (corresponding to the spherical GMM). The true number of parameters lies in between the ones selected by the BIC (underparameterized) and the AIC (overparameterized).

While the AIC and the BIC do not depend on the dimension $p$, alternative criteria such as the AICc~\citep{hurvich_regression_1989}---a small-sample version of the AIC---do. Some additional empirical results about the impact of the model selection criterion on the number of parameters of the PSA models are given in the appendix of~\cite{szwagier_curse_2025}. The AICc tends to select even more parsimonious models than the BIC for small-to-moderate sample sizes.

\subsection{Number of components $C$}\label{subsec:C}

In contrast to many works from the GMM literature that reduce the number of mixture parameters by reducing the number of components $C$, our work controls parsimony through the covariance structure of the mixture components. While $C$ has a \textit{linear} influence on the number of parameters, the eigenvalue multiplicities $\gamma_{ck}$ have a \textit{quadratic} influence on the number of parameters (see Equation~\eqref{eq:PSA_kappa}). Therefore, in high dimensions, we believe that it is much more interesting to reduce the number of parameters via the eigenvalue multiplicities than via the number of components (unless $C$ is very large).

Except for the image denoising experiments in Section~\ref{sec:denoising}, this paper voluntarily involves experiments where the number of components $C$ is well specified---i.e., equal to the true number of clusters. This choice is motivated by the willingness to highlight the impact of the covariance structure on the learned mixtures---that is the main message of the paper---rather than the impact of the number of clusters.

\begin{figure}
    \centering
    \includegraphics[width=\linewidth]{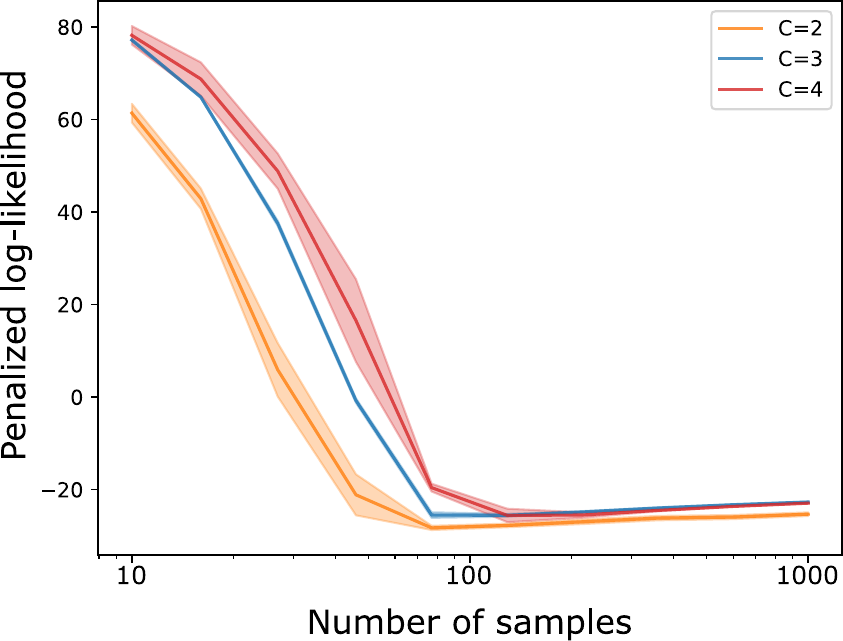}
    \caption{Influence of the number of components $C$ on the penalized log-likelihood, under the MPSA-U model. The true number of clusters is $C=3$.}
    \label{fig:C}
\end{figure}

In the current version of the code, the number of components $C$ can be updated during learning via several simple methodologies, for instance removing empty clusters, or the ones with less than a prespecified number of points. $C$ can also be estimated \textit{a posteriori}, as commonly done, via model selection criteria such as the BIC (see, for instance, \citet[Section~2.6]{bouveyron_model-based_2019}). As an illustration, we compare in Figure~\ref{fig:C} the penalized log-likelihood of an MPSA-U model with $C=2$, $C=3$ and $C=4$, under the setting of the previous experiment---with three clusters.
As we can see, the underfitted model ($C=2$) always has a significantly smaller penalized log-likelihood than the other models. Surprisingly, the overfitted model ($C=4$) achieves a significantly larger penalized log-likelihood than the true one ($C=3$) for $n \lessapprox 100$. Looking more closely at the estimated parameters, it seems that under the overfitted model, two clusters are well estimated, while the third one is split in two. This overparameterization naturally yields a larger log-likelihood, which is apparently not compensated by the penalty on the number of parameters---notably since the two overfitting components have lower number of parameters than the true one. Therefore, penalized log-likelihood criteria such as the BIC may have some limitations in selecting the true number of components, when the number of samples is small. For $n \gtrapprox 100$, the true model ($C=3$) gets a significantly larger penalized log-likelihood and is therefore selected over the overfitted one ($C=4$), testifying of the BIC's asymptotic consistency.

Beyond the two simple ideas proposed above (removing unpopulated clusters and using model selection criteria), we believe that the choice of the number of clusters could be thoroughly addressed via Dirichlet-type priors on the mixture proportions. Such priors are commonly used in the context of overfitted mixture models (see, notably, the classical works of~\citet{figueiredo_unsupervised_2002,rousseau_asymptotic_2011,malsiner-walli_model-based_2016}). They enable to automatically remove underpopulated clusters within the M-step, without losing the theoretical monotonicity guarantees of the EM algorithm.

\subsection{Patch size $s$ and number of components $C$ in denoising}\label{subsec:denoising_influence}
Except from Table~\ref{tab:denoising}, the main text's denoising experiments do not assess the impact of hyperparameters that are unrelated to the covariance structure on the results. This is the goal of this subsection.

\begin{figure}
    \centering
    \includegraphics[width=\linewidth]{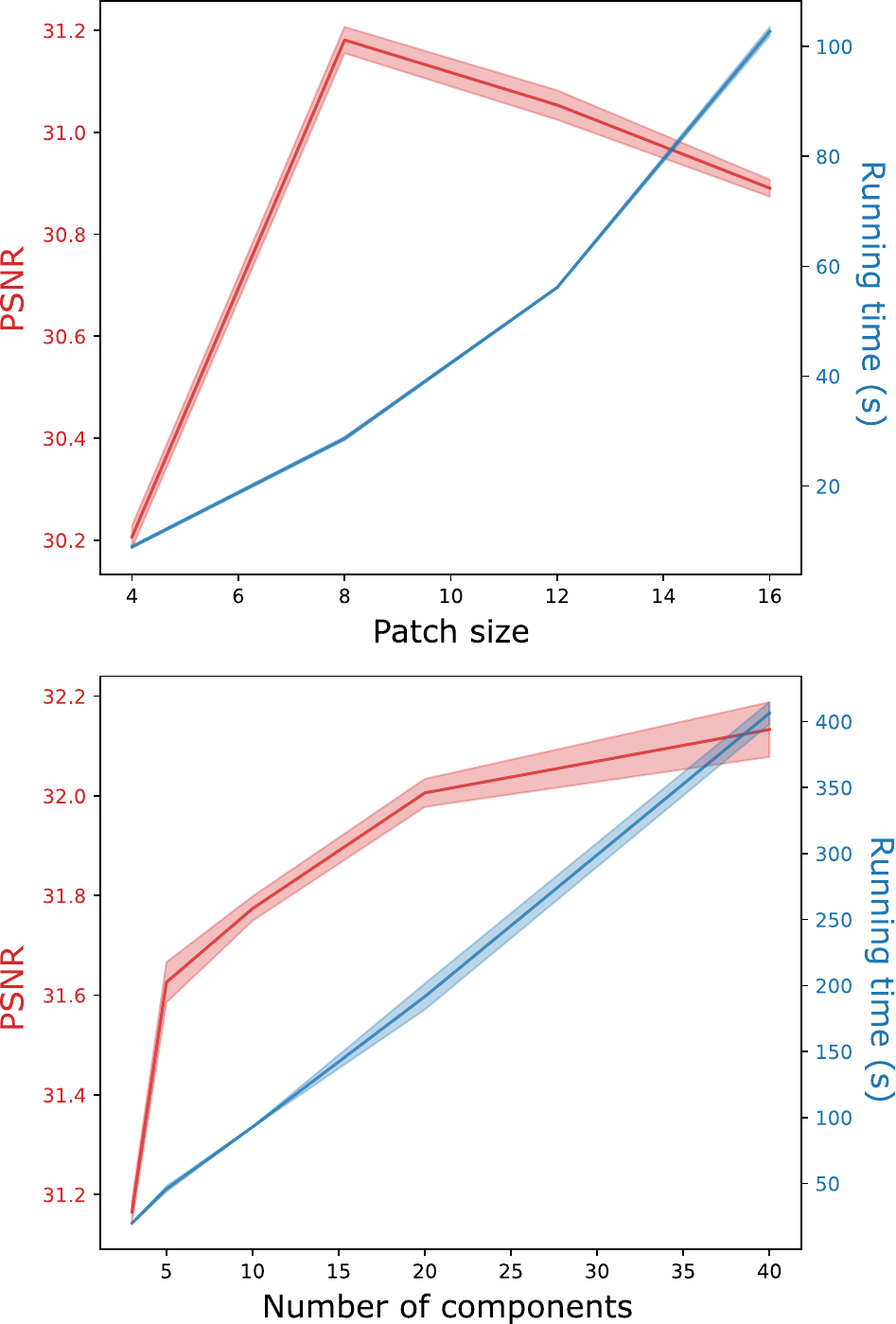}
    \caption{Influence of the patch size $s$ (top) and the number of components $C$ (bottom) on the denoising quality (PSNR) and running time, under the MPSA-U model.}
    \label{fig:denoising_influence}
\end{figure}

The size of the patches naturally influences the denoising results. Intuitively, if the patches are too small, then they all look similar, while if they are too large, then they all look different---especially in textured zones. This has notable repercussions on the denoising performances, since the investigated methods precisely rely on a faithful modeling of the groups of similar patches extracted from one image (see Section~\ref{subsec:denoising_literature}). Figure~\ref{fig:denoising_influence} reports the influence of the patch size on the quality of denoising (PSNR). We can see that it first increases with the patch size, up until approximately $8 \times 8$ (which is the size chosen in the experiments of Section~\ref{sec:denoising}) before decreasing. We can also see that the computing time naturally increases with the patch size, since the dimension of the dataset ($s^2$) increases quadratically with the patch size $s$. Therefore, beyond the denoising PSNR, the running time is another metric to take into account for choosing the patch size.

Similarly, the number of components $C$ does influence the denoising results. Intuitively, the goodness-of-fit increases with the number of components $C$. If the latter is too small, then we are in a situation of underfitting; the noise is overestimated and the denoising leads to a blurry image. In contrast, if the number of components is too large, then we are in a situation of overfitting; the noise is underestimated and the denoising does not modify much the input patch, resulting in a sharp yet noisy image; the limit case being for $C=n$, where each point is in its own cluster, the estimated noise is zero and the denoising leaves the noisy patches unchanged. These intuitions are (partially) confirmed empirically in Figure~\ref{fig:denoising_influence}, which reports the influence of the number of clusters on the PSNR and running times.\footnote{The decrease in denoising PSNR for large $C$ has been observed in~\citet[Section~4.1]{houdard_high-dimensional_2018}, but it appears to require an extremely large number of components to be significant, therefore we chose not to report it in Figure~\ref{fig:denoising_influence}.} Note that the running time increases linearly with $C$, therefore it might impact the practical choice of $C$ beyond the denoising performance. Notably, MPSA-U already reaches satisfying PSNRs for $C=5$.

\subsection{Truncated SVD}\label{subsec:TSVD}

The code for MPSA inference has been partially optimized in order to run in practical time. However, it could largely enjoy more engineering, especially for very large dimensions. One potential improvement is the truncated SVD (TSVD) evoked in Section~\ref{subsec:complexity}.

The current M-step for MPSA inference implies a full EVD of the responsibility-weighted covariance matrices $S_c$. Indeed, all the \textit{eigenvalues} are needed to select the appropriate eigenvalue multiplicities. However, not all the \textit{eigenvectors} are needed, but just the leading ones (as explained in Section~\ref{subsec:complexity}). Hence, an idea can be to first compute all the eigenvalues (\textit{without} the associated eigenvectors), then choose the appropriate eigenvalue multiplicities $\hat\gamma_c = (\hat\gamma_{c1}, \dots, \hat\gamma_{c d_c})$ and finally compute the $\hat q_c = p - \hat\gamma_{c d_c}$ leading eigenvectors (associated to the $\hat q_c$ largest eigenvalues) via truncated SVD. While such a trick should not change the outcome in theory---with respect to a full EVD---it actually does in practice, since the truncated SVD relies on different solvers from the classical EVD. More details can be found in the documentation of the scipy package~\citep{virtanen_scipy_2020}: see \href{https://docs.scipy.org/doc/scipy/reference/generated/scipy.linalg.eigh.html}{\texttt{scipy.linalg.eigh}} for the eigenvalue decomposition and \href{https://docs.scipy.org/doc/scipy/reference/generated/scipy.sparse.linalg.svds.html}{\texttt{scipy.sparse.linalg.svds}} for the truncated SVD. The randomized SVD of~\citet{halko_finding_2011} is another popular solver.

Another simple idea is to impose an upper bound on the intrinsic dimensions of the mixture components (i.e. $\hat q_c \leq q_\mathrm{max} \, \forall c$). In that case, one can directly run a truncated SVD with $q_\mathrm{max}$ components without the need to compute all the eigenvalues a priori. This approach can be expected to be more efficient than the previous one, as it does not require to compute all the eigenvalues first. However, imposing such an upper bound on the intrinsic dimensions restricts the set of candidate models and may therefore lead to suboptimal results.

\begin{figure}
    \centering
    \includegraphics[width=\linewidth]{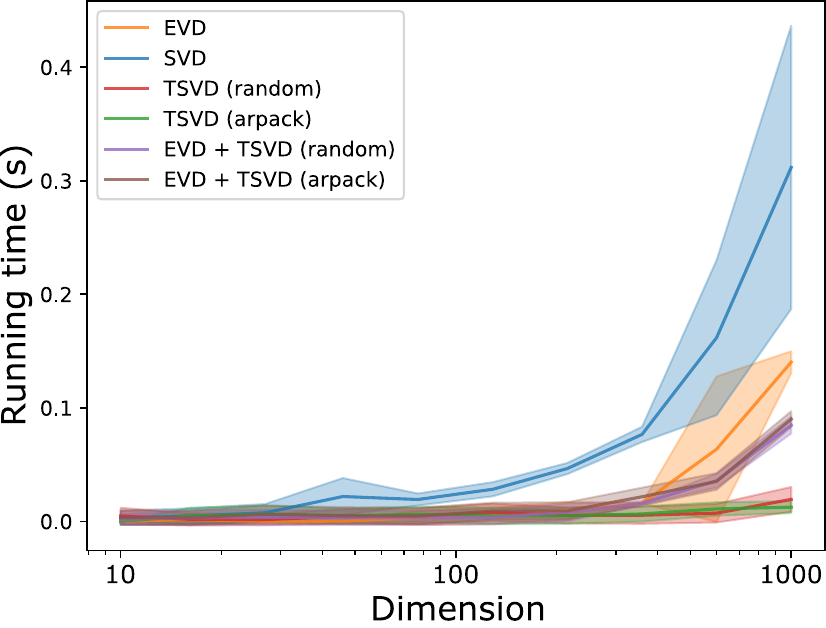}
    \caption{Comparison of the running times for different eigenvalue decomposition methodologies. EVD: full eigenvalue decomposition of the covariance matrix. SVD: full singular value decomposition of the data matrix. TSVD: truncated SVD of the data matrix, with five components (random: randomized SVD from~\cite{halko_finding_2011}; arpack: scipy solver which uses an Arnoldi iteration). EVD + TSVD: eigenvalue decomposition of the covariance matrix (\textit{without} the eigenvectors) followed by TSVD with five components.}
    \label{fig:MPSA_times_TSVD}
\end{figure}

To compare these different approaches, we sample $n=1000$ points from a multivariate Gaussian distribution with zero mean and covariance $\diag{10, 8, 6, 4, 2, 0.1^{p-5}}$, for varying $p$. We display the running times of the different approaches in Figure~\ref{fig:MPSA_times_TSVD}.
We see that the full SVD on the data matrix is overall slower than the other methodologies, including the full EVD on the covariance matrix. For very large dimensions ($p \gtrapprox 1000$), the methods involving first an eigenvalue decomposition without the eigenvectors and second a truncated SVD become significantly faster than the full EVD. The methods directly imposing an upper bound on the intrinsic dimensions are, for their part, significantly faster than the previous ones. Hence, we recommend integrating TSVD-based approaches in very large dimensions and directly imposing an upper bound on the intrinsic dimensions if speed and parsimony matter more than precise inference.

\subsection{Benchmark}
In our benchmark experiments, we mostly focus on comparing our MPSA models to the classical full and spherical GMMs---and to the less-classical HDDC~\citep{bouveyron_high-dimensional_2007} and HDMI~\citep{houdard_high-dimensional_2018}. This seemingly restrictive choice is actually well motivated: we want to assess the impact of equalizing close eigenvalues in covariance matrices on mixture modeling. We do not aim to achieve state-of-the-art results on density estimation, clustering and denoising tasks. Therefore, we believe that the best way to convey the main message of the paper (that is the interest of equalizing close eigenvalues in high-dimensional, low-sample settings) is to compare our models to baseline models that are particular cases of ours, and whose distinctive feature is the covariance structure \textit{alone}. Indeed, if we compared our models to other baseline models (for density estimation, clustering or denoising), then the difference in performance could be explained by other features than the covariance structure, which would drown the core message of the paper.

\subsection{Additional details}\label{subsec:details}
This subsection provides more details about the experiments. 
The code is written in Python 3.9 and run on an \textit{Intel® Core™ i7-10850H} CPU with 16 GB of RAM. The code for the EM algorithm is a modification of \href{https://github.com/mfauvel/HDDA}{Mathieu Fauvel's code} (license: GPL-3.0), itself inspired from the \textsf{R} package of~\cite{berge_hdclassif_2012} (license: GPL-2). 
Some parts of the code related to PSA inference and model selection were inspired from the code of~\cite{szwagier_curse_2025} (license: MIT). The code for the denoising pipeline is a modification of \href{https://github.com/ahoudard/HDMI}{Antoine Houdard's code} (license: MIT).

The used packages are the following: 
scikit-learn~\citep{pedregosa_scikit-learn_2011} (license: BSD 3), 
numpy~\citep{harris_array_2020} (license: BSD), 
scipy~\citep{virtanen_scipy_2020} (license: BSD 3), 
matplotlib~\citep{hunter_matplotlib_2007} (license: \href{https://matplotlib.org/stable/project/license.html}{custom}), 
\href{https://github.com/uci-ml-repo/ucimlrepo}{ucimlrepo} (license: MIT), scikit-image~\citep{walt_scikit-image_2014} (license: BSD 3). The code for sampling from a skew-normal distribution comes from  \href{https://gregorygundersen.com/blog/2020/12/29/multivariate-skew-normal/}{Gregory Gundersen's blog post} (license: not specified).

Some datasets come from the \href{https://archive.ics.uci.edu}{UCI Machine Learning repository}: wine~\citep{aeberhard_wine_1991} (license: CC BY 4.0), breast~\citep{wolberg_breast_1995} (license: CC BY 4.0) and ionosphere~\citep{sigillito_ionosphere_1989} (license: CC BY 4.0). The Simpson image, which is notably used in~\citep{houdard_high-dimensional_2018} comes from the \href{https://houdard.wp.imt.fr/files/2020/11/simpson_512.png}{website page} of Antoine Houdard (license: not specified).

\end{appendices}

\bibliography{sn-bibliography}

\end{document}